%% file: ms.tex
\documentclass{article}
\pdfoutput=1

\usepackage[numbers]{natbib}

\usepackage[preprint]{neurips_2019}

\usepackage[utf8]{inputenc} % allow utf-8 input
\usepackage[T1]{fontenc}    % use 8-bit T1 fonts
\usepackage{hyperref}       % hyperlinks
\usepackage{url}            % simple URL typesetting
\usepackage{booktabs}       % professional-quality tables
\usepackage{amsfonts}       % blackboard math symbols
\usepackage{nicefrac}       % compact symbols for 1/2, etc.
\usepackage{microtype}      % microtypography

\usepackage{algorithm,algorithmic}
\usepackage{amsmath}
\usepackage[pdftex]{graphicx}
\usepackage{caption}
\usepackage{enumitem}
\usepackage{multirow}
\usepackage{listings}
\usepackage{hyperref}

\title{Neural Consciousness Flow}

\author{
  Xiaoran Xu$^1$,\; Wei Feng$^1$,\; Zhiqing Sun$^2$,\; Zhi-Hong Deng$^2$ \\
  $^1$Hulu LLC, Beijing, China \\
  \texttt{\{xiaoran.xu, wei.feng\}@hulu.com} \\
  $^2$Peking University, Beijing, China \\
  \texttt{\{1500012783, zhdeng\}@pku.edu.cn} \\
}

\begin{document}

\maketitle

\input{p0_abstract}
\input{p1_introduction}

\input{p2_relatedwork}
\input{p3_model}
\input{p4_experiments}
\input{p5_conclusion}

\bibliography{reference}{}
\bibliographystyle{unsrt}

\newpage
\input{appendix}

\end{document}

%% file: p0_abstract.tex
\begin{abstract}

The ability of reasoning beyond data fitting is substantial to deep learning systems in order to make a leap forward towards artificial general intelligence. A lot of efforts have been made to model neural-based reasoning as an iterative decision-making process based on recurrent networks and reinforcement learning. Instead, inspired by \textit{the consciousness prior} proposed by Yoshua Bengio \cite{Bengio2017TheCP}, we explore reasoning with the notion of attentive awareness from a cognitive perspective, and formulate it in the form of attentive message passing on graphs, called \textbf{neural consciousness flow (NeuCFlow)}. Aiming to bridge the gap between deep learning systems and reasoning, we propose an attentive computation framework with a three-layer architecture, which consists of an \textit{unconsciousness flow layer}, a \textit{consciousness flow layer}, and an \textit{attention flow layer}. We implement the NeuCFlow model with graph neural networks (GNNs) and conditional transition matrices. Our attentive computation greatly reduces the complexity of vanilla GNN-based methods, capable of running on large-scale graphs. We validate our model for knowledge graph reasoning by solving a series of knowledge base completion (KBC) tasks. The experimental results show NeuCFlow significantly outperforms previous state-of-the-art KBC methods, including the embedding-based and the path-based. The reproducible code can be found by the link\footnote{https://github.com/netpaladinx/NeuCFlow} below.

\end{abstract}

%% file: p1_introduction.tex
\section{Introduction}

To discover the mystery of consciousness, several competing theories \cite{Dehaene1998ANM,Tononi2016IntegratedIT,Rosenthal2008HigherorderTO,van2004higher} have been proposed by neuroscientists. Despite their contradictory claims, they share a common notion that consciousness is a cognitive state of experiencing one's own existence, i.e. the state of awareness. Here, we do not refer to those elusive and mysterious meanings attributed to the word "consciousness". Instead, we focus on the basic idea, \textit{awareness} or \textit{attentive awareness}, to derive a neural network-based attentive computation framework on graphs, attempting to mimic the phenomenon of consciousness to some extent. 

The first work to bring the idea of attentive awareness into deep learning models, as far as we know, is Yoshua Bengio's consciousness prior \cite{Bengio2017TheCP}. He points out the process of disentangling higher-level abstract factors from full underlying representation and forming a low-dimensional combination of a few selected factors or concepts to constitute a conscious thought. Bengio emphasizes the role of attention mechanism in expressing awareness, which helps focus on a few elements of state representation at a given moment and combining them to make a statement, an action or policy. Two recurrent neural networks (RNNs), the representation RNN and the consciousness RNN, are used to summarize the current and recent past information and encode two types of state, the unconscious state denoted by a full high-dimensional vector before applying attention, and the conscious state by a derived low-dimensional vector after applying attention.

Inspired by the consciousness prior, we develop an attentive message passing mechanism. We model query-dependent states as motivation to drive iterative sparse access to an underlying large graph and navigate information flow via a few nodes to reach a target. Instead of using RNNs, we use two GNNs \cite{Scarselli2009TheGN,Battaglia2018RelationalIB} with node state representations. Nodes sense nearby topological structures by exchanging messages with neighbors, and then use aggregated information to update their states. However, the standard message passing runs globally and uniformly. Messages gathered by a node can come from possibly everywhere and get further entangled by aggregation operations. Therefore, we need to draw a query-dependent or context-aware local subgraph to guide message passing. Nodes within such a subgraph are densely connected, forming a community to further exchange and share information, reaching some resonance, and making subsequent decisions collectively to expand the subgraph and navigate information flow. To support such attentive information flow, we design an \textit{attention flow layer} above two GNNs. One GNN uses the standard message passing over a full graph, called \textit{unconsciousness flow layer}, while the other GNN runs on a subgraph built by attention flow, called \textit{consciousness flow layer}. These three flow layers constitute our attentive computation framework.

We realize the connection between attentive awareness and reasoning. A reasoning process is understood as a sequence of obvious or interpretable steps, either deductive, inductive, or abductive, to derive a less obvious conclusion. From the aspect of awareness, reasoning requires computation to be self-attentive or self-aware during processing in a way different from fitting by a black box. Therefore, interpretability must be one of the properties of reasoning. Taking KBC tasks as an example, many embedding-based models \cite{Bordes2013TranslatingEF,Yang2015EmbeddingEA,Dettmers2018Convolutional2K,Trouillon2016ComplexEF,Sun2018RotatEKG,Lacroix2018CanonicalTD} can do a really good job in link prediction, but lacking interpretation makes it hard to argue for their reasoning ability. People who aim at knowledge graph reasoning mainly focus on the path-based models using RL \cite{Lao2011RandomWI,Xiong2017DeepPathAR,Das2018GoFA,Shen2018MWalkLT} or logic-like methods \cite{Cohen2016TensorLogAD,Yang2017DifferentiableLO} to explicitly model a reasoning process to provide interpretations beyond predictions. Here, instead, we apply a flow-based attention mechanism, proposed in \cite{Xu2018ModelingAF}, as an alternative to RL for learning composition structure. In a manner of flowing, attention can propagate to cover a broader scope and increase the chance to hit a target. It maintains an end-to-end differentiable style, contrary to the way RL agents learn to choose a discrete action.

Other crucial properties of reasoning include relational inductive biases and iterative processing. Therefore, GNNs \cite{Scarselli2009TheGN,Battaglia2018RelationalIB} are a better choice compared to RNNs for encoding structured knowledge explicitly. Compared with the majority of previous GNN literature, focusing on the computation side, making neural-based architectures more composable and complex, we put a cognitive insight into it under the notion of attentive awareness. Specifically, we design an attention flow layer to chain attention operations directly with transition matrices, parallel to the message-passing pipeline to get less entangled with representation computation. This gives our model the ability to select edges step by step during computation and attend to a query-dependent subgraph, making a sharper prediction due to the disentanglement. These extracted subgraphs can reduce the computation cost greatly. In practice, we find our model can be applied to very large graphs with millions of nodes, such as the YAGO3-10 dataset, even running on a single laptop.

Our contributions are three-fold: (1) We propose an attentive computation framework on graphs, combining GNNs' representation power with explicit reasoning pattern, motivated by the cognitive notion of attentive awareness. (2) We exploit query-dependent subgraph structure, extracted by an attention flow mechanism, to address two shortcomings of most GNN implementations: the complexity and the non-context-aware aggregation schema. (3) We design a specific architecture for KBC tasks and demonstrate our model's strong reasoning capability compared to the state of the art, showing that a compact query-dependent subgraph is better than a path as a reasoning pattern.

%% file: p2_relatedwork.tex
\section{Related Work}

\begin{figure*}[t]
    \centering
    \includegraphics[width=0.7\textwidth]{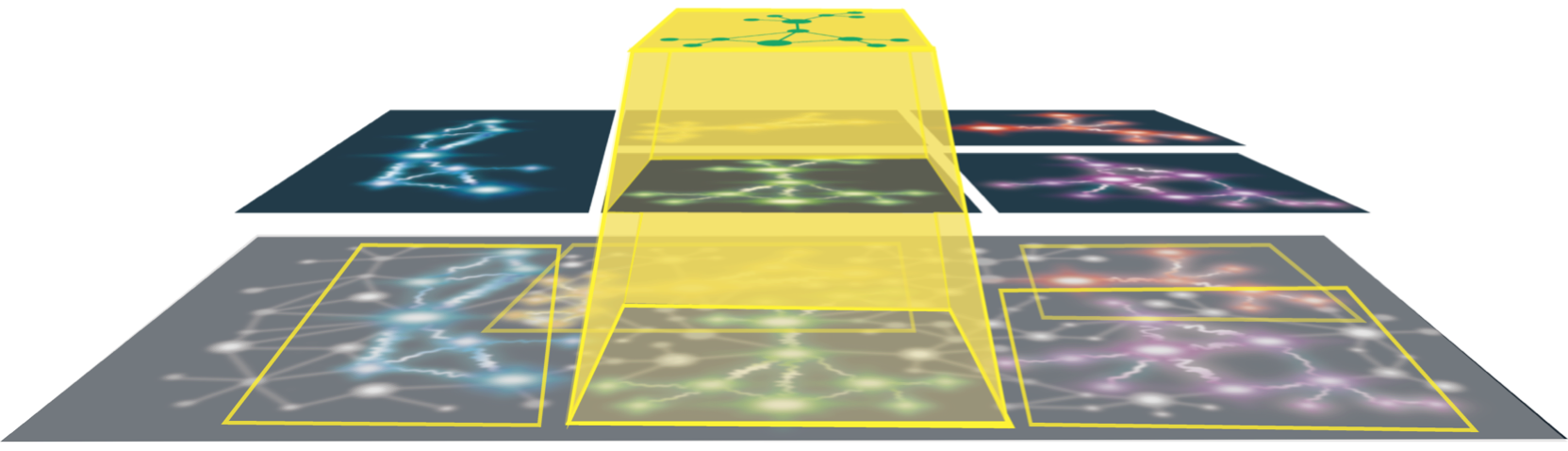}
    \caption{Illustration for the three-layer attentive computation framework. The bottom is a unified \textit{unconsciousness flow layer}, the middle contains small disentangled subgraphs to run attentive message passing separately, constituting a \textit{consciousness flow layer}, and the top is an \textit{attention flow layer} for extracting local subgraph structures.}
    \label{fig:neucflow_illustration}
    \vspace{-10pt}
\end{figure*}

\textbf{KBC and knowledge graph reasoning.} Early work for KBC, including TransE \cite{Bordes2013TranslatingEF} and its analogues \cite{Wang2014KnowledgeGE,Lin2015LearningEA,Ji2015KnowledgeGE}, DistMult \cite{Yang2015EmbeddingEA}, ConvE \cite{Dettmers2018Convolutional2K} and ComplEx \cite{Trouillon2016ComplexEF}, focuses on learning embeddings of entities and relations. Some recent work of this line \cite{Sun2018RotatEKG,Lacroix2018CanonicalTD} achieves high accuracy, yet unable to explicitly deal with compositional relationships that is crucial for reasoning. Another line aims to learn inference paths \cite{Lao2011RandomWI,Gardner2014IncorporatingVS,Guu2015TraversingKG,Lin2015ModelingRP,Toutanova2016CompositionalLO,Das2017ChainsOR} for knowledge graph reasoning, such as DeepPath \cite{Xiong2017DeepPathAR}, MINERVA \cite{Das2018GoFA}, and M-Walk \cite{Shen2018MWalkLT}, using RL to learn multi-hop relational paths over a graph towards a target given a query. However, these approaches, based on policy gradients or Monte Carlo tree search, often suffer from low sample efficiency and sparse rewards, requiring a large number of rollouts or running many simulations, and also the sophisticated reward function design. Other efforts include learning soft logical rules \cite{Cohen2016TensorLogAD,Yang2017DifferentiableLO} or compostional programs \cite{Liang2016NeuralSM} to reason over knowledge graphs.

\textbf{Relational reasoning by GNNs and attention mechanisms.} Relational reasoning is regarded as the key component of humans' capacity for combinatorial generalization, taking the form of entity- and relation-centric organization to reason about the composition structure of the world \cite{Craik1952TheNO,Anderson1982AcquisitionOC,Gentner1997StructureMI,Hummel2003AST,Lake2017BuildingMT}. A multitude of recent implementations \cite{Battaglia2018RelationalIB} encode relational inductive biases into neural networks to exploit graph-structured representation, including graph convolution networks (GCNs) \cite{Bruna2014SpectralNA,Henaff2015DeepCN,Duvenaud2015ConvolutionalNO,Kearnes2016MolecularGC,Defferrard2016ConvolutionalNN,Niepert2016LearningCN,Kipf2017SemiSupervisedCW,Bronstein2017GeometricDL} and graph neural networks \cite{Scarselli2009TheGN,Li2016GatedGS,Santoro2017ASN,Battaglia2016InteractionNF,Gilmer2017NeuralMP}, and overcome the difficulty to achieve relational reasoning for traditional deep learning models. These approaches have been widely applied to accomplishing real-world reasoning tasks (such as physical reasoning \cite{Battaglia2016InteractionNF,Chang2017ACO,Kipf2018NeuralRI,SanchezGonzalez2018GraphNA,Hamrick2018RelationalIB,Watters2017VisualIN}, visual reasoning \cite{Santoro2017ASN,Watters2017VisualIN,Raposo2017DiscoveringOA,Wang2018NonlocalNN,Chen2018IterativeVR}, textual reasoning \cite{Santoro2017ASN,Santoro2018RelationalRN,Palm2018RecurrentRN}, knowledge graph reasoning \cite{Kipf2017SemiSupervisedCW,OoroRubio2017RepresentationLF,Hamaguchi2017KnowledgeTF}, multiagent relationship reasoning \cite{Sukhbaatar2016LearningMC,Hoshen2017VAINAM}, and chemical reasoning \cite{Gilmer2017NeuralMP}), solving algorithmic problems (such as program verification \cite{Li2016GatedGS,Allamanis2018LearningTR}, combinatorial optimization \cite{Bello2017NeuralCO,Nowak2017ANO,Khalil2017LearningCO}, state transitions \cite{Johnson2017LearningGS}, and bollean satisfiability \cite{Selsam2018LearningAS}), or facilitating reinforcement learning with the structured reasoning or planning ability \cite{Hamrick2017MetacontrolFA,Pascanu2017LearningMP,SanchezGonzalez2018GraphNA,Hamrick2018RelationalIB,Wang2018NerveNetLS,Zambaldi2018RelationalDR,Toyer2018ActionSN}. Variants of GNN architectures have been developed with different focuses. Relation networks \cite{Santoro2017ASN} use a simple but effective neural module to equip deep learning models with the relational reasoning ability, and its recurrent versions \cite{Santoro2018RelationalRN,Palm2018RecurrentRN} do multi-step relational inference for long periods; Interaction networks \cite{Battaglia2016InteractionNF} provide a general-purpose learnable physics engine, and two of its variants are visual interaction networks \cite{Watters2017VisualIN} learning directly from raw visual data, and vertex attention interaction networks \cite{Hoshen2017VAINAM} with an attention mechanism; Message passing neural networks \cite{Gilmer2017NeuralMP} unify various GCNs and GCNs into a general message passing formalism by analogy to the one in graphical models.

Despite the strong representation power of GNNs, recent work points out its drawbacks that limit its capability. The vanilla message passing or neighborhood aggregation schema cannot adapt to strongly diverse local subgraph structure, causing performance degeneration when applying a deeper version or running more iterations \cite{Xu2018RepresentationLO}, since a walk of more steps might drift away from local neighborhood with information washed out via averaging. It is suggested that covariance rather than invariance to permutations of nodes and edges is preferable \cite{Kondor2018CovariantCN}, since being fully invariant by summing or averaging messages may worsen the representation power, lacking steerability. In this context, our model expresses permutation invariance under a constrained compositional transformation according to the group of possible permutations within each extracted query-dependent subgraph rather than the underlying full graph. Another drawback is the heavy computation complexity. GNNs are notorious for its poor scalability due to its quadratic complexity in the number of nodes when graphs are fully connected. Even scaling linearly with the number of edges by exploiting structure sparsity can still cause trouble on very large graphs, making selective or attentive computation on graphs so desirable.

Neighborhood attention operation can alleviate some limitation on GNNs' representation power by specifying different weights to different nodes or nodes' features \cite{Velickovic2018GraphAN,Hoshen2017VAINAM,Wang2018NonlocalNN,Kool2018AttentionSY}. These approaches often use multi-head self-attention to focus on specific interactions with neighbors when aggregating messages, inspired by \cite{Bahdanau2015NeuralMT,Lin2017ASS,Vaswani2017AttentionIA} originally for capturing long range dependencies. We notice that most graph-based attention mechanisms attend over neighborhood in a single-hop fashion, and \cite{Hoshen2017VAINAM} claims that the multi-hop architecture does not help in experiments, though they expect multiple hops to offer the potential to model high-order interaction. However, a flow-based design of attention in \cite{Xu2018ModelingAF} shows a promising way to characterize long distance dependencies over graphs, breaking the isolation of attention operations and stringing them in chronological order by transition matrices, like the spread of a random walk, parallel to the message-passing pipeline.

It is natural to extend relational reasoning to graph structure inference or graph generation, such as reasoning about a latent interaction graph explicitly to acquire knowledge of observed dynamics \cite{Kipf2018NeuralRI}, or learning generative models of graphs \cite{Li2018LearningDG,Cao2018MolGANAI,You2018GraphRNNGR,Bojchevski2018NetGANGG}. Soft plus hard attention mechanisms may be a better alternative to probabilistic models that is hard to train with latent discrete variables or might degenerate multi-step predictions due to the inaccuracy (biased gradients) of back-propagation.

%% file: p3_model.tex
\section{NeuCFlow Model}

\subsection{Attentive computation framework}

We extend Bengio's consciousness prior to graph-structured representation. Conscious thoughts are modeled by a few selected nodes and their edges, forming a context-aware subgraph, cohesive with sharper semantics, disentangled from the full graph. The underlying full graph forms the initial representation, entangled but rich, to help shape potential high-level subgraphs. We use attention flow to navigate conscious thoughts, capturing a step-by-step reasoning pattern. The attentive computation framework, as illustrated in Figure \ref{fig:neucflow_illustration}, consists of: (1) an \textit{unconsciousness flow (U-Flow) layer}, (2) a \textit{consciousness flow (C-Flow) layer}, and (3) an \textit{attention flow (A-Flow) layer}, with four guidelines to design a specific implementation as follows:
\begin{itemize}[wide=0pt, leftmargin=\dimexpr\labelwidth + 2\labelsep\relax]
    \item \textit{U-Flow} corresponds to a low-level computation graph for full state representation learning.
    \item \textit{C-Flow} contains high-level disentangled subgraphs for context-aware representation learning.
    \item \textit{A-Flow} is conditioned by both \textit{U-Flow} and \textit{C-Flow}, and also motivate \textit{C-Flow} but not \textit{U-Flow}.
    \item Information can be accessed by \textit{C-Flow} from \textit{U-Flow} with the help of \textit{A-Flow}.
\end{itemize}

\subsection{Model architecture design for knowledge graph reasoning}

\begin{figure*}[t]
  \centering
  \includegraphics[width=0.8\textwidth]{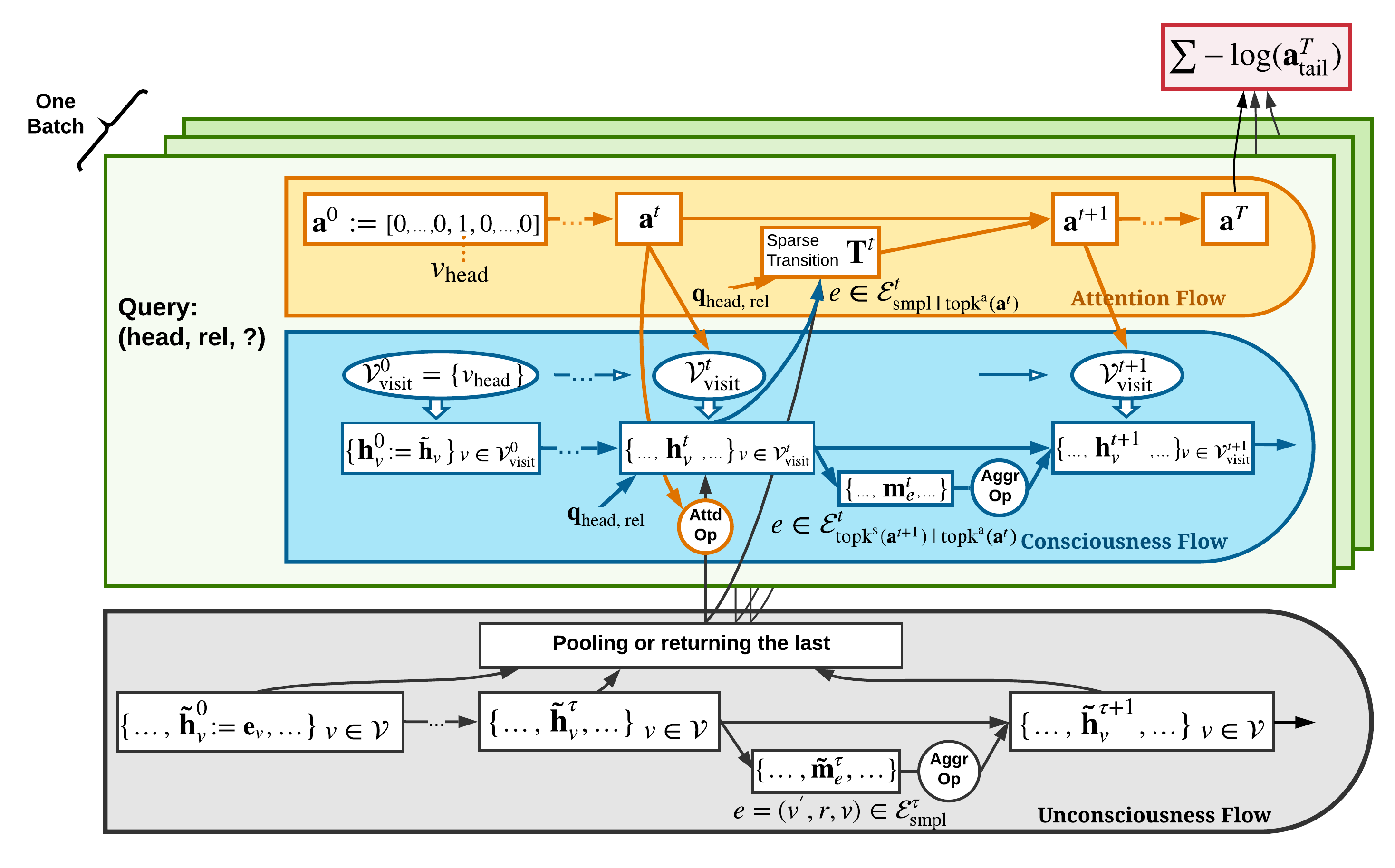}
  \caption{The neural consciousness flow architecture.}
  \label{fig:neucflow_architecture}
  \vspace{-10pt}
\end{figure*}

We choose KBC tasks to do KG reasoning. We let $\langle \mathcal{V}, \mathcal{E} \rangle$ denote a KG where $\mathcal{V}$ is a set of nodes (or entities) and $\mathcal{E}$ is a set of edges (or relations). A KG is viewed as a directed graph with each edge represented by a triple $\langle head, rel, tail \rangle$, where $head$ is the head entity, $tail$ is the tail entity, and $rel$ is their relation type. The aim of a KBC task is to predict potential unknown links, i.e., which entity is likely to be the tail given a query $\langle head, rel, ?\rangle$ with the head and the relation type specified.

The model architecture has three core components as shown in Figure \ref{fig:neucflow_architecture}. We here use the term "component" instead of "layer" to differentiate our flow layers from the referring normally used in neural networks, as each flow layer is more like a block containing many neural network layers.

\textbf{U-Flow component.} We implement this component over the full graph using the standard message passing mechanism \cite{Gilmer2017NeuralMP}. If the graph has an extremely large number of edges, we sample a subset of edges, $\mathcal{E}^\tau_\mathrm{smpl} \subset \mathcal{E}$, randomly each step when running message passing. For each batch of input queries, we let the representation computed by the U-Flow component be shared across these different queries, which means U-Flow is query-independent, with its state representation tensors containing no batch dimension, so that its complexity does not scale with the batch size and the saved computation resources can be allocated to sampling more edges. In U-Flow, each node $v$ has a learnable embedding $\mathbf{e}_v$ and a dynamical state $\mathbf{\tilde{h}}_v^{\tau}$ for step $\tau$, called unconscious node states, where the initial $\mathbf{\tilde{h}}_v^0:=\mathbf{e}_v$ for all $v \in \mathcal{V}$. Each edge type $r$ also has a learnable embedding $\mathbf{e}_r$, and edge $\langle v',r,v \rangle$ can produce a message, denoted by $\mathbf{\tilde{m}}_{\langle v',r,v \rangle}^{\tau}$, at step ${\tau}$. The U-Flow component includes:
\begin{itemize}[wide=0pt, leftmargin=\dimexpr\labelwidth + 2\labelsep\relax]
    \item Message function: $\mathbf{\tilde{m}}_{\langle v',r,v \rangle}^{\tau} =\psi_\mathrm{unc}(\mathbf{\tilde{h}}_{v'}^\tau, \mathbf{e}_r, \mathbf{\tilde{h}}_{v}^\tau)$, where $\langle v',r,v \rangle \in \mathcal{E}_\mathrm{smpl}^\tau$.
    \item Message aggregation: $\mathbf{\tilde{\mu}}_v^{\tau}=\frac{1}{\sqrt{\tilde{N}_v^\tau}} \sum_{v',r} \mathbf{\tilde{m}}_{\langle v',r,v \rangle}^\tau$, where $\langle v',r,v \rangle \in \mathcal{E}_\mathrm{smpl}^\tau$.
    \item Node state update function: $\mathbf{\tilde{h}}_v^{\tau+1}=\mathbf{\tilde{h}}_v^{\tau} + \delta_\mathrm{unc}(\tilde{\mu}_v^\tau, \mathbf{\tilde{h}}_v^{\tau}, \mathbf{e}_v)$, where $ v \in \mathcal{V}$.
\end{itemize}
We compute messages only for the sampled edges, $\langle v',r,v \rangle \in \mathcal{E}_\mathrm{smpl}^\tau$, each step. Functions $\psi_\mathrm{unc}$ and $\delta_\mathrm{unc}$ are implemented by a two-layer MLP (using $\mathrm{leakyReLu}$ for the first layer and $\mathrm{tanh}$ for the second layer) with input arguments concatenated respectively. Messages are aggregated by dividing the sum by the square root of the number of sampled neighbors that send messages, preserving the scale of variance. We use a residual adding to update each node state instead of a GRU or a LSTM. After running U-Flow for $\mathcal{T}$ steps, we return a pooling result or simply the last, $\mathbf{\tilde{h}}_v:=\mathbf{\tilde{h}}_v^{\mathcal{T}}$, to feed into downstream components.

\begin{figure*}[t]
  \centering
  \includegraphics[width=0.7\textwidth]{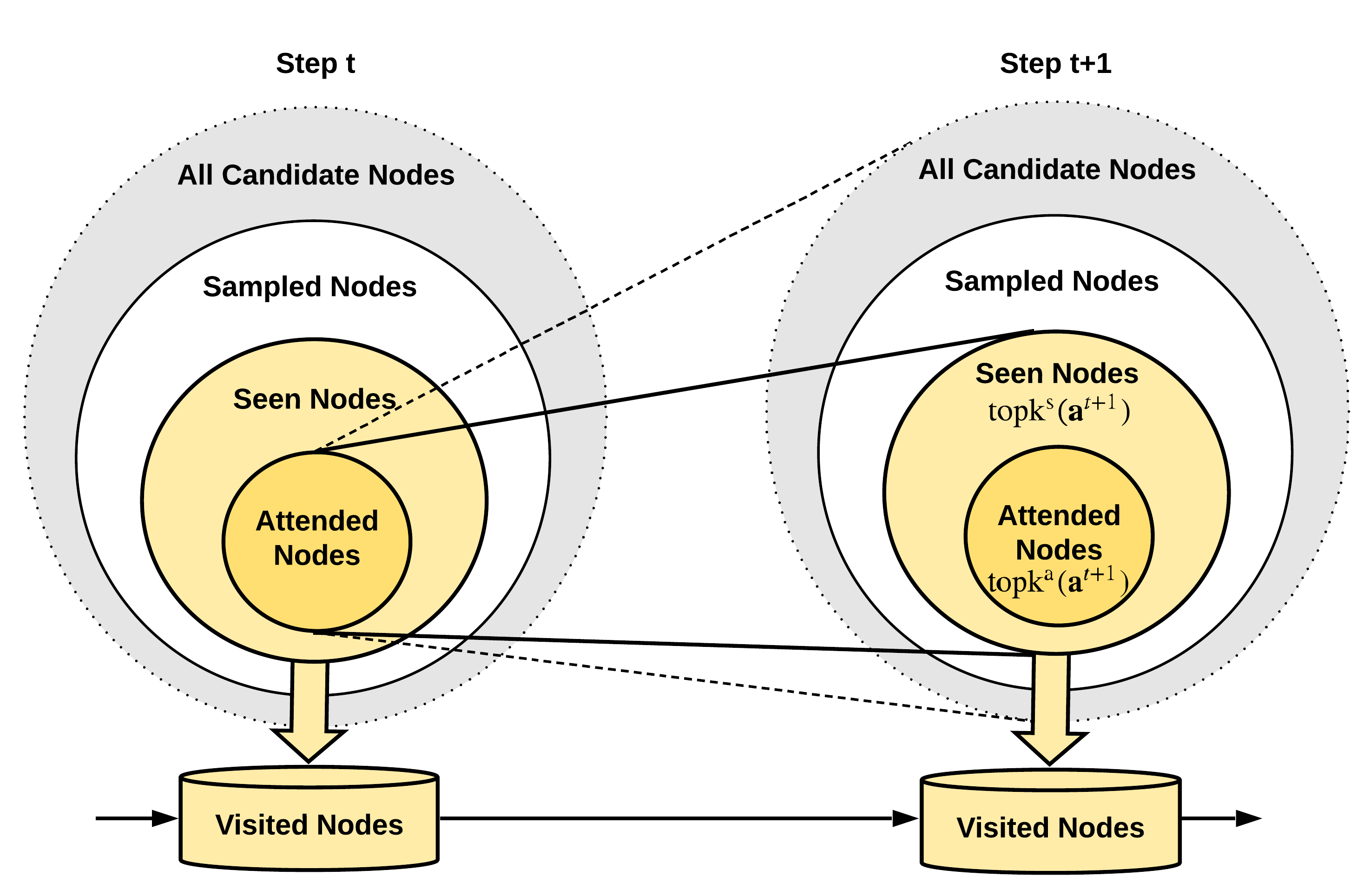}
  \caption{The iterative sampling-attending procedure for attentive complexity reduction, balancing the coverage as well as the complexity.}
  \label{fig:sampling_attending}
  \vspace{-10pt}
\end{figure*}

\textbf{C-Flow component.} C-Flow is query-dependent, which means that conscious node states, denoted by $\mathbf{h}_{v}^t$, have a batch dimension representing different input queries, making the complexity scale with the batch size. However, as C-Flow uses attentive message passing, running on small local subgraphs each conditioned by a query, we leverage the sparsity to record $\mathbf{h}_{v}^t$ only for the visited nodes $v \in \mathcal{V}_\mathrm{visit}^t$. For example, when $t=0$, for query $\langle head, rel, ? \rangle$, we start from node $head$, with $\mathcal{V}_\mathrm{visit}^0 = \{ v_{head} \}$ being a singleton, and thus record $\mathbf{h}_{v_{head}}^0$ only. When computing messages, denoted by $\mathbf{m}_{\langle v',r,v \rangle}^t$, in C-Flow, we use a sampling-attending procedure, explained in Section \ref{sec:sampling_attending}, to further control the number of computed edges. The C-Flow component has:
\begin{itemize}[wide=0pt, leftmargin=\dimexpr\labelwidth + 2\labelsep\relax]
    \item Message function: $\mathbf{m}_{\langle v',r,v \rangle}^t=\psi_\mathrm{con}(\mathbf{h}_{v'}^t, \mathbf{c}_r, \mathbf{h}_{v}^t)$, where $\langle v',r,v \rangle \in \mathcal{E}_{\mathrm{topk^s}(\mathbf{a}^{t+1})\,|\,\mathrm{topk^a}(\mathbf{a}^t)}^t$, and $\mathbf{c}_{r}=[\mathbf{e}_r, \mathbf{q}_{head}, \mathbf{q}_{rel}]$.
    \item Message aggregation: $\mathbf{\mu}_v^t=\frac{1}{\sqrt{N_v^t}} \sum_{v',r} \mathbf{m}_{\langle v',r,v \rangle}^t$, where $\langle v',r,v \rangle \in \mathcal{E}_{\mathrm{topk^s}(\mathbf{a}^{t+1})\,|\,\mathrm{topk^a}(\mathbf{a}^t)}^t$.
    \item Node state attending function: $\tilde{\mathbf{\eta}}_v^{t+1} = a^{t+1}_v\mathbf{A}\cdot \tilde{\mathbf{h}}_v$, where $a^{t+1}_v=\mathbf{a}^{t+1}[v]$ and $v\in \mathcal{V}_\mathrm{visit}^{t+1}$.
    \item Node state update function: $\mathbf{h}_v^{t+1}=\mathbf{h}_v^{t} + \delta_\mathrm{con}(\mu_v^t, \mathbf{h}_v^{t}, \mathbf{c}_{v}^{t+1})$, where $\mathbf{c}_{v}^{t+1}=[\tilde{\mathbf{\eta}}_v^{t+1}, \mathbf{q}_{head}, \mathbf{q}_{rel}]$.
\end{itemize}
C-Flow and U-Flow share the embeddings $\mathbf{e}_r$. A query is represented by its head and relation embeddings, $\mathbf{q}_{head}:=\mathbf{e}_{head}$ and $\mathbf{q}_{rel}:=\mathbf{e}_{rel}$, participating in computing messages and updating node states. We here select a subset of edges, $\mathcal{E}_{\mathrm{topk^s}(\mathbf{a}^{t+1})\,|\,\mathrm{topk^a}(\mathbf{a}^t)}^t$, rather than sampling, according to edges between the attended nodes at step $t$ and the seen nodes at step $t+1$, defined in Section \ref{sec:sampling_attending}, as shown in Figure \ref{fig:sampling_attending}. We introduce the node state attending function to pass an unconscious state $\tilde{\mathbf{h}}_v$ to C-Flow adjusted by a scalar attention $a^{t+1}_v$ and a learnable matrix $\mathbf{A}$. We initialize $\mathbf{h}_v^0:=\tilde{\mathbf{h}}_v$ for $v \in \mathcal{V}_\mathrm{visit}^0$, treating the rest as zero states.

\textbf{A-Flow component.} Attention flow is represented by a series of probability distributions changing across steps, denoted as $\mathbf{a}^t, t=1,2\ldots,T$. The initial distribution $\mathbf{a}^0$ is a one-hot vector with $\mathbf{a}^0[v_{head}] = 1$. To spread attention, we need to compute transition matrices $\mathbf{T}^t$ each step. Given that A-Flow is conditioned by both U-Flow and C-Flow, we model the transition from $v'$ to $v$ by two types of interaction: conscious-to-conscious, $\mathbf{h}_{v'}^{t} \sim \mathbf{h}_{v}^{t}$, and conscious-to-unconscious, $\mathbf{h}_{v'}^{t} \sim \mathbf{\tilde{h}}_{v}$. The former favors previously visited nodes, while the latter is useful to attend to unseen nodes.
$$\mathbf{T}^t[:, v']=\mathrm{softmax}_{v\in \mathcal{N}_{v'}^t}\big(\sum\nolimits_r \alpha_\mathrm{cc}(\mathbf{h}_{v'}^{t}, \mathbf{c}_r, \mathbf{h}_{v}^{t}) + \sum\nolimits_r \alpha_\mathrm{cu}(\mathbf{h}_{v'}^{t}, \mathbf{c}_r, \mathbf{\tilde{h}}_{v})\big)$$
where $\alpha_\mathrm{cc} = \mathrm{MLP}(\mathbf{h}_{v'}^{t}, \mathbf{c}_r)^\mathrm{T} \mathbf{\Theta}_\mathrm{cc} \mathrm{MLP}(\mathbf{h}_{v}^{t}, \mathbf{c}_r)$ and $\alpha_\mathrm{cu} = \mathrm{MLP}(\mathbf{h}_{v'}^{t}, \mathbf{c}_r)^\mathrm{T} \mathbf{\Theta}_\mathrm{cu} \mathrm{MLP}(\mathbf{\tilde{h}}_{v}, \mathbf{c}_r)$, and $\mathbf{\Theta}_\mathrm{cc}$ and $\mathbf{\Theta}_\mathrm{cu}$ are two learnable matrices. Each MLP uses one single layer with the $\mathrm{leakyReLu}$ activation. To reduce the complexity for computing $\mathbf{T}^t$, we select attended nodes, $v' \in \mathrm{topk^a}(\mathbf{a}^t)$, which is the set of nodes with the k-largest attention, and then sample $v$ from $v'$ neighbors as next nodes. Then, we compute a sparse $\mathbf{T}^t$ according to edges $\langle v',r,v \rangle \in \mathcal{E}_{\mathrm{smpl\,|\,topk^a}(\mathbf{a}^t)}$. Due to the fact that the attended nodes may not carry all attention, a small amount of attention can be lost during transition, causing the total amount to decrease. Therefore, we use a renormalized version, $\mathbf{a}^{t+1}=\mathbf{T}^t\mathbf{a}^t / \|\mathbf{T}^t\mathbf{a}^t\|$. We use the final attention on the tail as the probability for prediction to compute the training objective, as shown in Figure \ref{fig:neucflow_architecture}.

\subsection{Complexity reduction by iterative sampling and attending}
\label{sec:sampling_attending}

Previously, we use edge sampling, in a globally and uniformly random manner, to address the complexity issue in U-Flow, where we are not concerned about the batch size. Here, we need to confront the complexity that scales with the batch size in C-Flow. Suppose that we run a normal message passing for $T$ steps on a KG with $|\mathcal{V}|$ nodes and $|\mathcal{E}|$ edges for a batch of $N$ queries. Then, the complexity is $\mathcal{O}(NTD(|\mathcal{V}|+|\mathcal{E}|))$ where $D$ represents the number of representation dimensions. The complexity can be reduced to $\mathcal{O}(NTD(|\mathcal{V}|+|\mathcal{E}_\mathrm{smpl}|))$ by using edges sampling. $T$ is a small positive integer, often less than $10$. $D$ is normally between $50$ and $200$, and being too small for $D$ would lead to underfitting. In U-Flow, we have $N = 1$, while in C-Flow, let us say $N = 100$. Then, to maintain the same complexity as U-Flow, we have to reduce the sampling rate by a factor of $100$ on each query. However, the U-Flow's edge sampling procedure is for the full graph, and it is inappropriate to  apply to C-Flow on each query due to the reduced sample rate. Also, when $|\mathcal{V}|$ becomes as large as $|\mathcal{E}_\mathrm{smpl}|$, we also need to consider decreasing $|\mathcal{V}|$. 

Good news is that C-Flow deals with a local subgraph for each query so that we only record a few selected nodes, called \textit{visited nodes}, denoted by $\mathcal{V}_\mathrm{visit}^t$. We can see that $|\mathcal{V}_\mathrm{visit}^t|$ is much less than $|\mathcal{V}|$. The initial $\mathcal{V}_\mathrm{visit}^0$, when $t=0$, contains only one node $v_{head}$, and then $\mathcal{V}_\mathrm{visit}^t$ is enlarged each step by adding new nodes during spreading. When propagating messages, we only care about the one-step neighborhood each step. However, the spreading goes so rapidly that after only a few steps it covers almost all nodes, causing the number of computed edges to increase dramatically. The key to address the problem is that we need to constrain the scope of nodes we jump from each step, i.e., the core nodes that determine where we can go based on where we depart from. We call them \textit{attended nodes}, which are in charge of the attending-from horizon, selected by $\mathrm{topk}^\mathrm{a}(\mathbf{a}^t)$ based on the current attention $\mathbf{a}^t$. Given the set of attended nodes, we still need edge sampling over their neighborhoods in case of a hub node of extremely high degree. Here, we face a tricky problem that is to make a trade-off between the coverage and the complexity when sampling over the neighborhoods. Also, we need to well maintain these coherent context-aware node states and avoid possible noises or drifting away caused by sampling neighbors randomly. Therefore, we introduce an attending-to horizon inside the sampling horizon. We compute A-Flow over the sampling horizon with a smaller dimension to compute the attention, exchanged for sampling more neighbors to increase the coverage. Based on the newly computed attention $\mathbf{a}^{t+1}$, we select a smaller subset of nodes, $\mathrm{topk}^\mathrm{s}(\mathbf{a}^{t+1})$, to receive messages in C-Flow, called \textit{seen nodes}, in charge of the attending-to horizon. The next attending-from horizon is chosen by $\mathrm{topk}^\mathrm{a}(\mathbf{a}^{t+1}) \subset \mathrm{topk}^\mathrm{s}(\mathbf{a}^{t+1})$, a sub-horizon of the current attending-to horizon. All seen and attended nodes are stored as visited nodes along steps. We illustrate this sampling-attending procedure in Figure \ref{fig:sampling_attending}.

To compute our reduced complexity, we let $N_\mathrm{e}$ be the maximum number of sampled edges per attended node per step, $N_\mathrm{s}$ the maximum number of seen nodes per step, and $N_\mathrm{a}$ the maximum number of attended nodes per step. We also denote the dimension number used in A-Flow as $D_a$. For one batch, the complexity of C-Flow is $\mathcal{O}(NTD(N_\mathrm{a} + N_\mathrm{s} + N_\mathrm{a}N_\mathrm{s}))$ for the worst case, where attended and seen nodes are fully connected, and $\mathcal{O}(NTD\cdot c(N_\mathrm{a} + N_\mathrm{s}))$ in most cases, where $c$ is a small constant. The complexity of A-Flow is $\mathcal{O}(NT D_a N_\mathrm{a} N_\mathrm{e})$ where $D_a$ is much smaller than $D$.

%% file: p4_experiments.tex
\section{Experiments}

\newcommand{\spa}{\space\space\space\space\space\space\space}

\subsection{Datasets and experimental settings}

\setlength{\tabcolsep}{4pt}
\begin{table}[t]
  \caption{Statistics of the six KG datasets. A KG is built on all training triples including their inverse triples. Note that we do not count the inverse triples in FB15K, FB15K-237, WN18, WN18RR, and YAGO3-10 as shown below to be consistent with the statistics reported in other papers, though we include them in the training, validation and test set. PME (tr) means the proportion of multi-edge triples in train; PME (te) means the proportion of multi-edge triples in test; AvgD (te) means the average length of shortest paths connecting each head-tail pair in test.}
  \label{tab:datasets_stats}
  \centering
  \begin{tabular}{l|ccccc|ccc}
    \noalign{\hrule height 1.5pt}
    Dataset & \#Entities & \#Rels & \#Train & \#Valid & \#Test & PME (tr) & PME (te) & AvgD (te)  \\
    \hline
    FB15K & 14,951 & 1,345 & 483,142 & 50,000 & 59,071 & 81.2\% & 80.9\% & 1.22 \\
    FB15K-237 & 14,541 & 237 & 272,115 & 17,535 & 20,466 & 38.0\% & \textbf{0\%} & 2.25 \\
    WN18 & 40,943 & 18 & 141,442 & 5,000 & 5,000 & 93.1\% & 94.0\% & 1.18 \\
    WN18RR & 40,943 & 11 & 86,835 & 3,034 & 3,134 & 34.5\% & 35.0\% & 2.87 \\
    NELL995 & 74,536 & 200 & 149,678 & 543 & 2,818 & 100\% & 41.0\% & 2.06 \\
    YAGO3-10 & 123,188 & 37 & 1,079,040 & 5,000 & 5,000 & 56.4\% & 56.0\% & 1.75  \\
  \noalign{\hrule height 1.5pt}
  \end{tabular}
  \vspace{-5pt}
\end{table}

\setlength{\tabcolsep}{3.5pt}
\begin{table}[t]
  \caption{Comparison results on the FB15K-237 and WN18RR datasets. Results of [$\spadesuit$] are taken from \cite{Nguyen2018ANE}, results of [$\clubsuit$] from \cite{Dettmers2018Convolutional2K}, results of [$\heartsuit$] from \cite{Shen2018MWalkLT}, results of [$\diamondsuit$] from \cite{Sun2018RotatEKG}, and results of [$\triangle$] from \cite{Das2018GoFA}. Some collected results only have a metric score while some including ours take the form of "mean (standard deviation)".}
  \label{tab:comparison_results_fb237_wn18rr}
  \centering
  \begin{tabular}{l|cccc|cccc}
    \noalign{\hrule height 1.5pt}
    & \multicolumn{4}{c|}{\textbf{FB15K-237}} & \multicolumn{4}{|c}{\textbf{WN18RR}} \\
    Metric ($\%$) & H@1 & H@3 & H@10 & MRR & H@1 & H@3 & H@10 & MRR \\
    \hline
    \hline
    TransE [$\spadesuit$] & - & - & 46.5 & 29.4 & - & - & 50.1 & 22.6 \\
    \hline
    DistMult [$\clubsuit$] & 15.5 \spa & 26.3 \spa & 41.9 & 24.1 \spa & 39 \spa & 44 \spa & 49 & 43 \spa \\
    DistMult [$\heartsuit$] & 20.6 (.4) & 31.8 (.2) & - & 29.0 (.2) & 38.4 (.4) & 42.4 (.3) & - & 41.3 (.3) \\
    \hline
    ComplEx [$\clubsuit$] & 15.8 \spa & 27.5 \spa & 42.8 & 24.7 \spa & 41 \spa & 46 \spa & 51 & 44 \spa \\
    ComplEx [$\heartsuit$] & 20.8 (.2) & 32.6 (.5) & - & 29.6 (.2) &  38.5 (.3) & 43.9 (.3) & - & 42.2 (.2)  \\
    \hline
    ConvE [$\clubsuit$] & 23.7 \spa & 35.6 \spa & 50.1 & 32.5 \spa & 40 \spa & 44 \spa & 52 & 43 \spa \\
    ConvE [$\heartsuit$] & 23.3 (.4) & 33.8 (.3) & - & 30.8 (.2) & 39.6 (.3) & 44.7 (.2) & - & 43.3 (.2) \\
    \hline
    RotatE [$\diamondsuit$] & 24.1 \spa & 37.5 \spa & \textbf{53.3} & 33.8 \spa & 42.8 \spa & 49.2 \spa & \textbf{57.1} & 47.6 \spa \\
    \hline
    \hline
    NeuralLP [$\heartsuit$] & 18.2 (.6)	& 27.2 (.3) & - & 24.9 (.2) & 37.2 (.1) & 43.4 (.1) & - & 43.5 (.1) \\
    \hline
    MINERVA [$\heartsuit$] & 14.1 (.2) & 23.2 (.4) & - & 20.5 (.3) & 35.1 (.1) & 44.5 (.4) & - & 40.9 (.1) \\
    MINERVA [$\triangle$] & - & - & 45.6 & - & 41.3 \spa & 45.6 \spa & 51.3 & - \\
    \hline
    M-Walk [$\heartsuit$] & 16.5 (.3) &	24.3 (.2) & - & 23.2 (.2) & 41.4 (.1) & 44.5 (.2) & - & 43.7 (.1) \\
    \hline
    \hline
    \textbf{NeuCFlow} & \textbf{28.6 (.1)} & \textbf{40.3 (.1)} & 53.0 (.3) & \textbf{36.9 (.1)} & \textbf{44.4 (.4)} & \textbf{49.7 (.8)} & 55.8 (.5) & \textbf{48.2 (.5)} \\
    \noalign{\hrule height 1.5pt}
  \end{tabular}
  \vspace{-5pt}
\end{table}

\textbf{Datasets.} We evaluate our model using six large KG datasets\footnote{https://github.com/netpaladinx/NeuCFlow/tree/master/data}: FB15K, FB15K-237, WN18, WN18RR, NELL995, and YAGO3-10. FB15K-237 \cite{Toutanova2015ObservedVL} is sampled from FB15K \cite{Bordes2013TranslatingEF} with redundant relations removed, and WN18RR \cite{Dettmers2018Convolutional2K} is a subset of WN18 \cite{Bordes2013TranslatingEF} removing triples that cause test leakage. Thus, they are both considered more challenging. NELL995 \cite{Xiong2017DeepPathAR} has separate datasets for 12 query relations each corresponding to a single-query-relation KBC task. YAGO3-10 \cite{Mahdisoltani2014YAGO3AK} contains the largest KG with millions of edges. Their statistics are shown in Table \ref{tab:datasets_stats}. We find some statistical differences between train and test. In a KG with all training triples as its edges, a triple $(head, rel, tail)$ is considered as a multi-edge triple if the KG contains other triples that also connect $head$ and $tail$ ignoring the direction. We notice that FB15K-237 is a special case compared with the others, as there are no edges in its KG directly linking any pair of $head$ and $tail$ in test. Therefore, when using training triples as queries to train our model, given a batch, for FB15K-237, we cut off from the KG all triples connecting the head-tail pairs in the given batch, ignoring relation types and edge directions, forcing the model to learn a composite reasoning pattern rather than a single-hop pattern, and for the rest datasets, we only remove the triples of this batch and their inverse from the KG before training on this batch.

\textbf{Experimental settings.} We use the same data split protocol as in many papers \cite{Dettmers2018Convolutional2K,Xiong2017DeepPathAR,Das2018GoFA}. We create a KG, a directed graph, consisting of all train triples and their inverse added for each dataset except NELL995, since it already includes reciprocal relations. Besides, every node in KGs has a self-loop edge to itself. We also add inverse relations into the validation and test set to evaluate the two directions. For evaluation metrics, we use HITS@1,3,10 and the mean reciprocal rank (MRR) in the filtered setting for FB15K-237, WN18RR, FB15K, WN18, and YAGO3-10, and use the mean average precision (MAP) for NELL995's single-query-relation KBC tasks. For NELL995, we follow the same evaluation procedure as in \cite{Xiong2017DeepPathAR, Das2018GoFA, Shen2018MWalkLT}, ranking the answer entities against the negative examples given in their experiments. We run our experiments using a 12G-memory GPU, TITAN X (Pascal), with Intel(R) Xeon(R) CPU E5-2670 v3 @ 2.30GHz. Our code is written in Python based on TensorFlow 2.0 and NumPy 1.16.

\subsection{Baselines and comparison results}

\setlength{\tabcolsep}{4pt}
\begin{table}[t]
  \caption{Comparison results on the FB15K and WN18 datasets. Results of [$\spadesuit$] are taken from \cite{Nickel2016HolographicEO}, results of [$\clubsuit$] are from \cite{Dettmers2018Convolutional2K}, results of [$\diamondsuit$] are from \cite{Sun2018RotatEKG}, and results of [$\heartsuit$] are from \cite{Yang2017DifferentiableLO}. Our results take the form of "mean (standard deviation)".}
  \label{tab:comparison_results_fb15k_wn18}
  \centering
  \begin{tabular}{l|cccc|cccc}
    \noalign{\hrule height 1.5pt}
    & \multicolumn{4}{c|}{\textbf{FB15K}} & \multicolumn{4}{|c}{\textbf{WN18}} \\
    Metric ($\%$) & H@1 & H@3 & H@10 & MRR & H@1 & H@3 & H@10 & MRR \\
    \hline
    \hline
    TransE [$\spadesuit$] & 29.7 & 57.8 & 74.9 & 46.3 & 11.3 & 88.8 & 94.3 & 49.5 \\
    \hline
    HolE [$\spadesuit$] & 40.2 & 61.3 & 73.9 & 52.4 & 93.0 & 94.5 & 94.9 & 93.8 \\
    \hline
    DistMult [$\clubsuit$] & 54.6 & 73.3 & 82.4 & 65.4 & 72.8 & 91.4 & 93.6 & 82.2 \\
    \hline
    ComplEx [$\clubsuit$] & 59.9 & 75.9 & 84.0 & 69.2 & 93.6 & 93.6 & 94.7 & 94.1 \\
    \hline
    ConvE [$\clubsuit$] & 55.8 & 72.3 & 83.1 & 65.7 & 93.5 & 94.6 & 95.6 & 94.3 \\
    \hline
    RotatE [$\diamondsuit$] & \textbf{74.6} & \textbf{83.0} & \textbf{88.4} & \textbf{79.7} & \textbf{94.4} & \textbf{95.2} & \textbf{95.9} & \textbf{94.9} \\
    \hline
    \hline
    NeuralLP [$\heartsuit$] & - & - & 83.7 & 76 & - & - & 94.5 & 94  \\
    \hline
    \hline
    \textbf{NeuCFlow} & 72.6 (.4) & 78.4 (.4) & 83.4 (.5) & 76.4 (.4) & 91.6 (.8) & 93.6 (.4) & 94.9 (.4) & 92.8 (.6) \\
    \noalign{\hrule height 1.5pt}
  \end{tabular}
  \vspace{-5pt}
\end{table}

\begin{table}[t]
  \caption{Comparison results on the YAGO3-10 dataset. Results of [$\spadesuit$] are taken from \cite{Dettmers2018Convolutional2K}, and results of [$\clubsuit$] are from \cite{Lacroix2018CanonicalTD}.}
  \label{tab:comparison_results_yago310}
  \centering
  \begin{tabular}{l|cccc}
    \noalign{\hrule height 1.5pt}
    & \multicolumn{4}{c}{\textbf{YAGO3-10}} \\
    Metric ($\%$) & H@1 & H@3 & H@10 & MRR \\
    \hline
    \hline
    DistMult [$\spadesuit$] & 24 & 38 & 54 & 34 \\
    \hline
    ComplEx [$\spadesuit$] & 26 & 40 & 55 & 36 \\
    \hline
    ConvE [$\spadesuit$] & 35 & 49 & 62 & 44 \\
    \hline
    ComplEx-N3 [$\clubsuit$] & - & - & \textbf{71} & \textbf{58} \\
    \hline
    \hline
    \textbf{NeuCFlow} & \textbf{48.4} & \textbf{59.5} & 67.9 & 55.3 \\
    \noalign{\hrule height 1.5pt}
  \end{tabular}
  \vspace{-5pt}
\end{table}

\begin{table}[t]
  \caption{Comparison results of MAP scores ($\%$) on NELL995's single-query-relation KBC tasks. We take our baselines' results from \cite{Shen2018MWalkLT}. All results take the form of "mean (standard deviation)" except for TransE and TransR.}
  \label{tab:comparison_results_nell995}
  \centering
  \begin{tabular}{lc|ccccc}
    \noalign{\hrule height 1.5pt}
    Tasks & \textbf{NeuCFlow} & M-Walk & MINERVA & DeepPath & TransE & TransR  \\
    \hline
    AthletePlaysForTeam & 83.9 (0.5) & \textbf{84.7 (1.3)} & 82.7 (0.8) & 72.1 (1.2) & 62.7 & 67.3 \\
    AthletePlaysInLeague & 97.5	(0.1) & \textbf{97.8 (0.2)} & 95.2 (0.8) & 92.7 (5.3) & 77.3 & 91.2 \\
    AthleteHomeStadium & \textbf{93.6 (0.1)} & 91.9 (0.1) & 92.8 (0.1) & 84.6 (0.8) & 71.8 & 72.2 \\
    AthletePlaysSport & \textbf{98.6 (0.0)} & 98.3 (0.1) &\textbf{ 98.6 (0.1)} & 91.7 (4.1) & 87.6 & 96.3 \\
    TeamPlayssport & \textbf{90.4 (0.4)} & 88.4 (1.8) & 87.5 (0.5) & 69.6 (6.7) & 76.1 & 81.4 \\
    OrgHeadQuarteredInCity & 94.7 (0.3) & \textbf{95.0 (0.7)} & 94.5 (0.3) & 79.0 (0.0) & 62.0 & 65.7 \\
    WorksFor & \textbf{86.8 (0.0)} & 84.2 (0.6) & 82.7 (0.5) & 69.9 (0.3) & 67.7 & 69.2 \\
    PersonBornInLocation & \textbf{84.1 (0.5)} & 81.2 (0.0) & 78.2 (0.0) & 75.5 (0.5) & 71.2 & 81.2 \\
    PersonLeadsOrg & 88.4 (0.1) & \textbf{88.8 (0.5)} & 83.0 (2.6) & 79.0 (1.0) & 75.1 & 77.2 \\
    OrgHiredPerson & 84.7 (0.8) & \textbf{88.8 (0.6)} & 87.0 (0.3) & 73.8 (1.9) & 71.9 & 73.7 \\
    AgentBelongsToOrg & \textbf{89.3 (1.2)} & - & - & - & - & -  \\				
    TeamPlaysInLeague & \textbf{97.2 (0.3)} & - & - & - & - & -  \\					
    \noalign{\hrule height 1.5pt}
  \end{tabular}
  \vspace{-5pt}
\end{table}

\textbf{Baselines.} We compare our model against embedding-based approaches, including TransE \cite{Bordes2013TranslatingEF}, TransR \cite{Lin2015LearningEA}, DistMult \cite{Yang2015EmbeddingEA}, ConvE \cite{Dettmers2018Convolutional2K}, ComplE \cite{Trouillon2016ComplexEF}, HolE \cite{Nickel2016HolographicEO}, RotatE \cite{Sun2018RotatEKG}, and ComplEx-N3 \cite{Lacroix2018CanonicalTD}, and path-based approaches that use RL methods, including DeepPath \cite{Xiong2017DeepPathAR}, MINERVA \cite{Das2018GoFA}, and M-Walk \cite{Shen2018MWalkLT}, and also that uses learned neural logic, NeuralLP \cite{Yang2017DifferentiableLO}. For all the baselines, we quote the results from the corresponding papers instead of rerunning them. For our method, we run the experiments three times in each hyperparameter setting on each dataset to report the means and standard deviations of the results. We put the details of our hyperparameter settings in the appendix.

\textbf{Comparison results and analysis.} We first report the comparison on FB15K-23 and WN18RR in Table \ref{tab:comparison_results_fb237_wn18rr}. NeuCFlow has a surprisingly good result, significantly outperforming all the compared methods in HITS@1,3 and MRR on both the two datasets. Compared to the best baseline, RotatE, published very recently, we only lose a few points in HITS@10 but gain a lot in HITS@1,3 and MRR. Based on the observation that NeuCFlow gains a larger amount of advantage when k in HITS@k gets smaller, we speculate that the reasoning ability acquired by NeuCFlow is to make a sharper prediction by exploiting graph-structured composition locally and conditionally, in contrast to embedding-based methods, which totally rely on vectorized representation. When a target becomes too vague to predict, reasoning may lose its great advantage, though still very competitive. However, path-based baselines, with a certain ability to do KG reasoning, perform worse than we expect. We argue that it is inappropriate to think of reasoning, a sequential decision process, as a sequence of nodes, i.e. a path, in KGs. The average length of the shortest paths between heads and tails in the test set in a KG, as shown in Table \ref{tab:datasets_stats}, suggests an extremely short path, making the motivation for using a path pattern almost pointless. The iterative reasoning pattern should be characterized in the form of dynamically varying local graph-structured patterns, holding a bunch of nodes resonating with each other to produce a decision collectively. Then, we run our model on larger KGs, including FB15K, WN18, and YAGO3-10, and summarize the comparison in Table \ref{tab:comparison_results_fb15k_wn18},\ref{tab:comparison_results_yago310}, where NeuCFlow beats most well-known baselines and achieves a very competitive position against the best state-of-the-art methods. Moreover, we summarize the comparison on NELL995's tasks in Table \ref{tab:comparison_results_nell995}. NeuCFlow performs the best on five tasks, also being very competitive against M-Walk, the best path-based method as far as we know, on the rest. We find no reporting on the last two tasks from the corresponding papers.

\subsection{Experimental analysis}

\begin{figure*}[t]
  \centering
  \includegraphics[width=\textwidth]{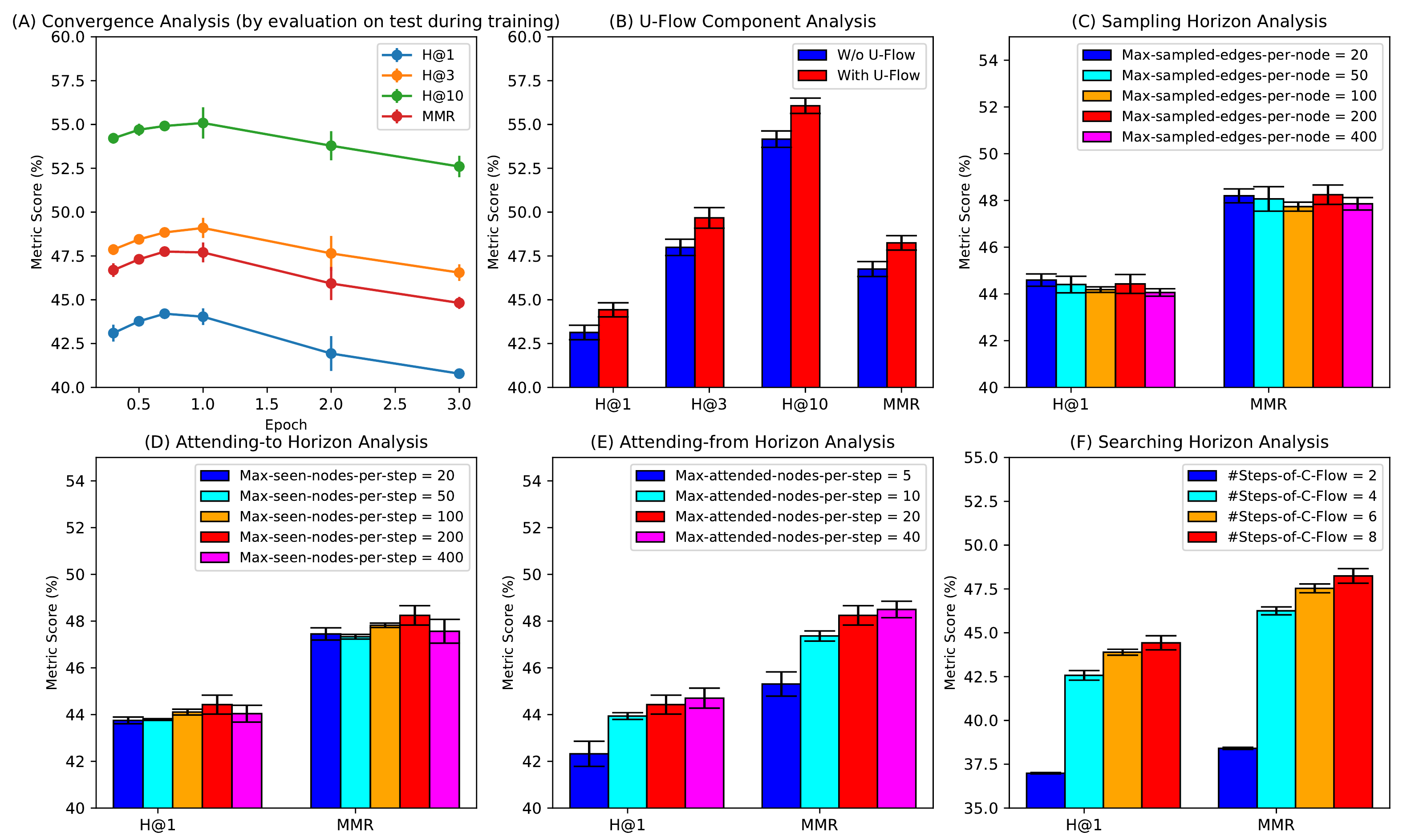}
  \caption{Experimental analysis on WN18RR: (A) During training we pick six model snapshots at time points of 0.3, 0.5, 0.7, 1, 2, and 3 epochs and evaluate them on test; (B) The \textit{w/o U-Flow} uses zero step to run U-Flow, while the \textit{with U-Flow} uses two steps; (C)-(F) are for the sampling, attending and searching horizon analysis based on the standard hyperparameter settings listed in the appendix. The experimental analysis charts on FB15K-237 can be found in the appendix.}
  \label{fig:wn18rr_analysis}
  \vspace{-5pt}
\end{figure*}

\textbf{Convergence analysis.} During training we find that NeuCFlow converges surprisingly fast. We may use half of training examples to get the model well trained and generalize it to the test, sometimes producing an even better metric score than trained for a full epoch, as shown in Figure \ref{fig:wn18rr_analysis}(A). Compared with the less expensive computation using embedding-based models, although our model takes a large number of edges to compute for each input query, consuming more time on one batch, it does not need a second epoch or even taking all training triples as queries in one epoch, thus saving a lot of training time. The reason may be that all queries are directly from the KG's edge set and some of them have probably been exploited to construct subgraphs for many times during the training of other queries, so that we might not have to train the model on each query explicitly as long as we have other ways to exploit them.

\textbf{Component analysis.} If we do not run U-Flow, then the unconscious state $\mathbf{\tilde{h}}_v$ is just the initial embedding of node $v$, and we can still run C-Flow as usual. We want to know whether the U-Flow component is actually useful. Considering that long-distance message passing might bring in less informative features, we compare running U-Flow for two steps against totally shutting it down. The result in Figure \ref{fig:wn18rr_analysis}(B) shows that U-Flow brings a small gain in each metric on WN18RR.

\textbf{Horizon analysis.} The sampling, attending and searching horizons determine how large area the flow can spread over. They impact the computation complexity as well as the performance of the model with different degrees depending on the properties of a dataset. Intuitively, enlarging the probe scope by sampling more, attending more, or searching longer, may increase the chance to hit a target. However, the experimental results in Figure \ref{fig:wn18rr_analysis}(C)(D) show that it is not always the case. In Figure \ref{fig:wn18rr_analysis}(E), we can see that increasing the maximum number of the attending-from nodes, i.e. attended nodes, per step is more important, but our GPU does not allow for a larger number to accommodate more intermediate data produced during computation, otherwise causing the error of \textit{ResourceExhaustedError}. Figure \ref{fig:wn18rr_analysis}(F) shows the step number of C-Flow cannot get too small as two.

\textbf{Attention flow analysis.} If attention flow can really capture the way we reason about the world, its process should be conducted in a diverging-converging thinking pattern. Intuitively, first, for the diverging thinking, we search and collect ideas as much as we can; then, for the converging thinking, we try to concentrate our thoughts on one point. To check whether the attention flow has such a pattern, we measure the average entropy of attention distributions varying along steps and also the proportion of attention concentrated at the top-1,3,5 attended nodes. As we expect, attention indeed is more focused at the final step as well as at the beginning.  

\begin{figure*}[t]
  \centering
  \includegraphics[width=\textwidth]{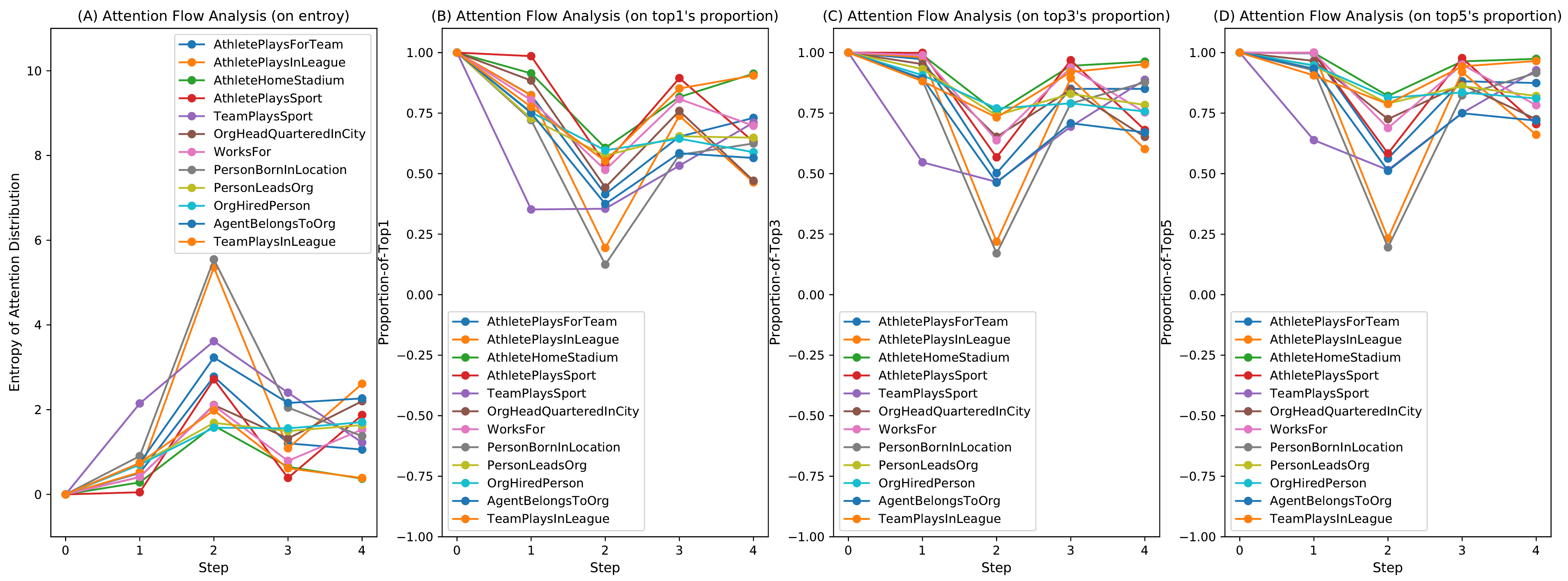}
  \caption{Analysis of attention flow on NELL995 tasks: (A) records how the average entropy of attention distributions varies along steps for each single-query-relation KBC task. (B)(C)(D) measure the changing of the proportion of attention concentrated at the top-1,3,5 attended nodes per step for each task.}
  \label{fig:nell995_analysis}
  \vspace{-5pt}
\end{figure*}

\begin{figure*}[t]
  \centering
  \includegraphics[width=\textwidth]{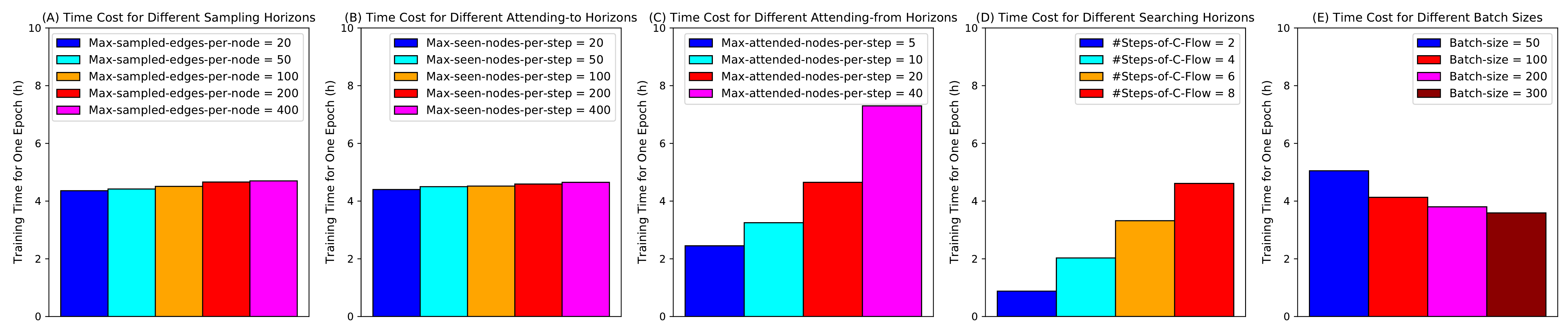}
  \caption{Analysis of time cost on WN18RR: (A)-(D) measure the training time for one epoch on different horizon settings corresponding to Figure \ref{fig:wn18rr_analysis}(C)-(F); (E) measures the training time for one epoch for different batch sizes using the same horizon setting, which is \textit{Max-sampled-edges-per-node}=20, \textit{Max-seen-nodes-per-step}=20, \textit{Max-attended-nodes-per-step}=20, and \textit{\#Steps-of-C-Flow}=8. The time cost analysis charts on FB15K-237 can be found in the appendix.}
  \label{fig:time_cost_wn18rr}
  \vspace{-5pt}
\end{figure*}

\textbf{Time cost analysis.} The time cost is affected not only by the scale of a dataset but also by the horizon setting. For each dataset, we list the training time for one epoch corresponding to the standard hyperparameter settings in the appendix. Note that there is always a trade-off between the complexity and the performance. We thus study whether we can reduce the time cost a lot at the price of sacrificing a little performance. We plot the one-epoch training time in Figure \ref{fig:time_cost_wn18rr}(A)-(D), using the same settings as we do in the horizon analysis. We can see that \textit{Max-attended-nodes-per-step} and \textit{\#Steps-of-C-Flow} affect the training time significantly while \textit{Max-sampled-edges-per-node} and \textit{Max-seen-nodes-per-step} affect very slightly. Therefore, we can use smaller \textit{Max-sampled-edges-per-node} and \textit{Max-seen-nodes-per-step} in order to gain a larger batch size, making the computation more efficiency as shown in Figure \ref{fig:time_cost_wn18rr}(E).

\subsection{Visualization}

To further demonstrate the reasoning ability acquired by our model, we show some visualization results of the extracted subgraphs on NELL995's test data for 12 separate tasks. We avoid using the training data in order to show the generalization of our model's learned reasoning ability on knowledge graphs. Here, we show the visualization result for the \textit{AthletePlaysForTeam} task. The rest can be found in the appendix.

\textbf{For the AthletePlaysForTeam task}

\begin{figure}[h]
  \hspace{-35pt}
  \includegraphics[width=1.2\textwidth]{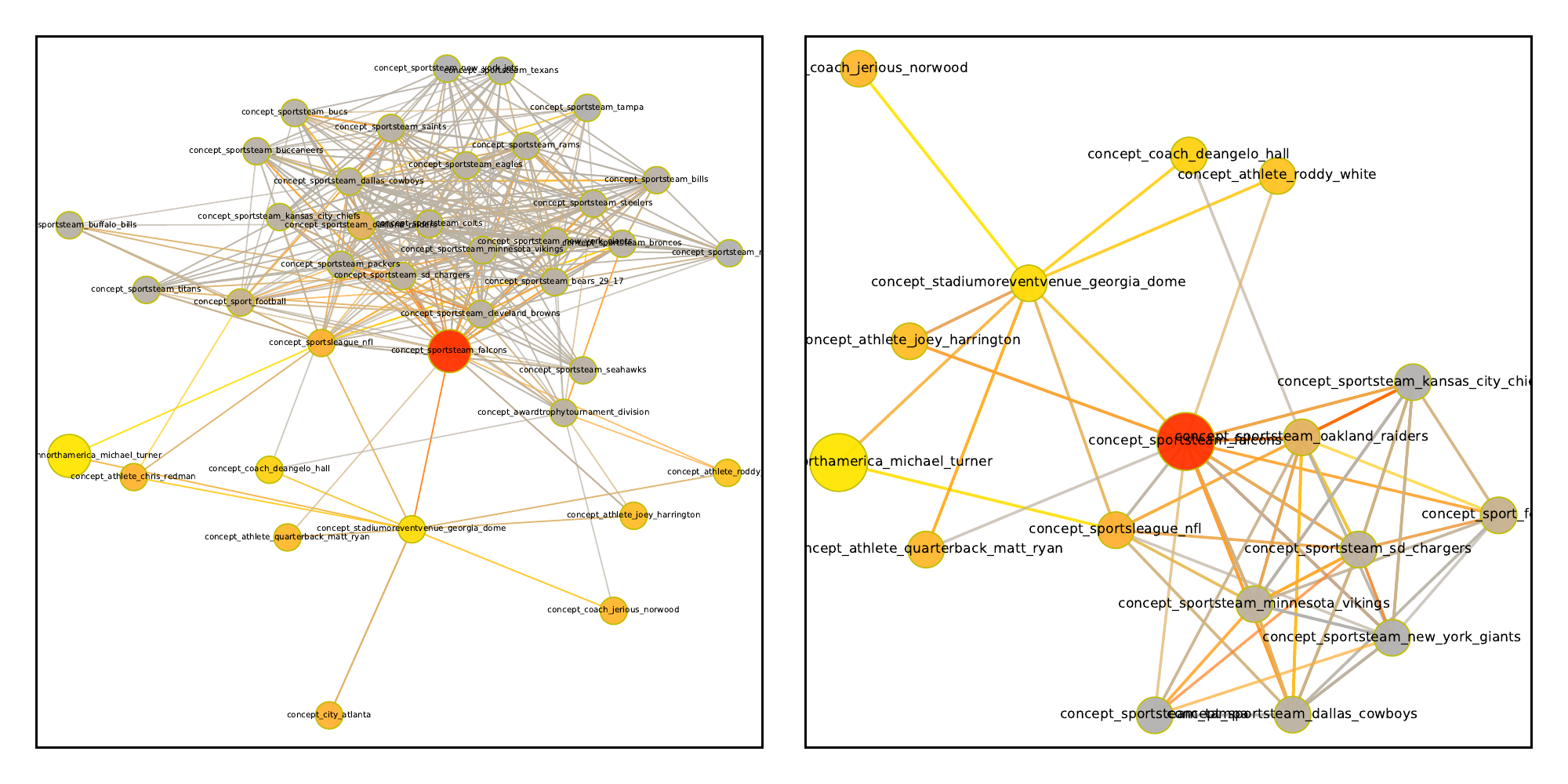}
  \caption{\textbf{AthletePlaysForTeam}. The head is \textit{concept\_personnorthamerica\_michael\_turner}, the query relation is \textit{concept:athleteplaysforteam}, and the desired tail is \textit{concept\_sportsteam\_falcons}. The left is a full subgraph derived with $max\_attended\_nodes\_per\_step=20$, and the right is a further extracted subgraph from the left based on attention. The big yellow node represents the head, and the big red node represents the tail. Colors indicate how important a node is attended to in a local subgraph. Grey means less important, yellow means it is more attended during the early steps, and red means it is more attended when getting close to the final step.}
  \label{fig:athleteplaysforteam}
  \vspace{-10pt}
\end{figure}

\begin{lstlisting}[basicstyle=\sffamily\tiny]
Query: (concept_personnorthamerica_michael_turner, concept:athleteplaysforteam, concept_sportsteam_falcons)

Selected key edges:
concept_personnorthamerica_michael_turner, concept:agentbelongstoorganization, concept_sportsleague_nfl
concept_personnorthamerica_michael_turner, concept:athletehomestadium, concept_stadiumoreventvenue_georgia_dome
concept_sportsleague_nfl, concept:agentcompeteswithagent, concept_sportsleague_nfl
concept_sportsleague_nfl, concept:agentcompeteswithagent_inv, concept_sportsleague_nfl
concept_sportsleague_nfl, concept:teamplaysinleague_inv, concept_sportsteam_sd_chargers
concept_sportsleague_nfl, concept:leaguestadiums, concept_stadiumoreventvenue_georgia_dome
concept_sportsleague_nfl, concept:teamplaysinleague_inv, concept_sportsteam_falcons
concept_sportsleague_nfl, concept:agentbelongstoorganization_inv, concept_personnorthamerica_michael_turner
concept_stadiumoreventvenue_georgia_dome, concept:leaguestadiums_inv, concept_sportsleague_nfl
concept_stadiumoreventvenue_georgia_dome, concept:teamhomestadium_inv, concept_sportsteam_falcons
concept_stadiumoreventvenue_georgia_dome, concept:athletehomestadium_inv, concept_athlete_joey_harrington
concept_stadiumoreventvenue_georgia_dome, concept:athletehomestadium_inv, concept_athlete_roddy_white
concept_stadiumoreventvenue_georgia_dome, concept:athletehomestadium_inv, concept_coach_deangelo_hall
concept_stadiumoreventvenue_georgia_dome, concept:athletehomestadium_inv, concept_personnorthamerica_michael_turner
concept_sportsleague_nfl, concept:subpartoforganization_inv, concept_sportsteam_oakland_raiders
concept_sportsteam_sd_chargers, concept:teamplaysinleague, concept_sportsleague_nfl
concept_sportsteam_sd_chargers, concept:teamplaysagainstteam, concept_sportsteam_falcons
concept_sportsteam_sd_chargers, concept:teamplaysagainstteam_inv, concept_sportsteam_falcons
concept_sportsteam_sd_chargers, concept:teamplaysagainstteam, concept_sportsteam_oakland_raiders
concept_sportsteam_sd_chargers, concept:teamplaysagainstteam_inv, concept_sportsteam_oakland_raiders
concept_sportsteam_falcons, concept:teamplaysinleague, concept_sportsleague_nfl
concept_sportsteam_falcons, concept:teamplaysagainstteam, concept_sportsteam_sd_chargers
concept_sportsteam_falcons, concept:teamplaysagainstteam_inv, concept_sportsteam_sd_chargers
concept_sportsteam_falcons, concept:teamhomestadium, concept_stadiumoreventvenue_georgia_dome
concept_sportsteam_falcons, concept:teamplaysagainstteam, concept_sportsteam_oakland_raiders
concept_sportsteam_falcons, concept:teamplaysagainstteam_inv, concept_sportsteam_oakland_raiders
concept_sportsteam_falcons, concept:athleteledsportsteam_inv, concept_athlete_joey_harrington
concept_athlete_joey_harrington, concept:athletehomestadium, concept_stadiumoreventvenue_georgia_dome
concept_athlete_joey_harrington, concept:athleteledsportsteam, concept_sportsteam_falcons
concept_athlete_joey_harrington, concept:athleteplaysforteam, concept_sportsteam_falcons
concept_athlete_roddy_white, concept:athletehomestadium, concept_stadiumoreventvenue_georgia_dome
concept_athlete_roddy_white, concept:athleteplaysforteam, concept_sportsteam_falcons
concept_coach_deangelo_hall, concept:athletehomestadium, concept_stadiumoreventvenue_georgia_dome
concept_coach_deangelo_hall, concept:athleteplaysforteam, concept_sportsteam_oakland_raiders
concept_sportsleague_nfl, concept:teamplaysinleague_inv, concept_sportsteam_new_york_giants
concept_sportsteam_sd_chargers, concept:teamplaysagainstteam_inv, concept_sportsteam_new_york_giants
concept_sportsteam_falcons, concept:teamplaysagainstteam, concept_sportsteam_new_york_giants
concept_sportsteam_falcons, concept:teamplaysagainstteam_inv, concept_sportsteam_new_york_giants
concept_sportsteam_oakland_raiders, concept:teamplaysagainstteam_inv, concept_sportsteam_new_york_giants
concept_sportsteam_oakland_raiders, concept:teamplaysagainstteam, concept_sportsteam_sd_chargers
concept_sportsteam_oakland_raiders, concept:teamplaysagainstteam_inv, concept_sportsteam_sd_chargers
concept_sportsteam_oakland_raiders, concept:teamplaysagainstteam, concept_sportsteam_falcons
concept_sportsteam_oakland_raiders, concept:teamplaysagainstteam_inv, concept_sportsteam_falcons
concept_sportsteam_oakland_raiders, concept:agentcompeteswithagent, concept_sportsteam_oakland_raiders
concept_sportsteam_oakland_raiders, concept:agentcompeteswithagent_inv, concept_sportsteam_oakland_raiders
concept_sportsteam_new_york_giants, concept:teamplaysagainstteam, concept_sportsteam_sd_chargers
concept_sportsteam_new_york_giants, concept:teamplaysagainstteam, concept_sportsteam_falcons
concept_sportsteam_new_york_giants, concept:teamplaysagainstteam_inv, concept_sportsteam_falcons
concept_sportsteam_new_york_giants, concept:teamplaysagainstteam, concept_sportsteam_oakland_raiders
\end{lstlisting}

In the above case, the query is (\textit{concept\_personnorthamerica\_michael\_turner}, \textit{concept:athleteplays-forteam}, ?) and the desired answer is \textit{concept\_sportsteam\_falcons}. From Figure \ref{fig:athleteplaysforteam}, we can see our model learns that (\textit{concept\_personnorthamerica\_michael\_turner}, \textit{concept:athletehomestadium}, \textit{concept\_stadiumoreventvenue\_georgia\_dome}) and (\textit{concept\_stadiumoreventvenue\_georgia\_dome}, \textit{concept:teamhomestadium\_inv}, \textit{concept\_sportsteam\_falcons}) are two important facts to support the answer of \textit{concept\_sportsteam\_falcons}. Besides, other facts, such as (\textit{concept\_athlete\_joey\_harrington}, \textit{concept:athletehomestadium}, \textit{concept\_stadiumoreventvenue\_georgia\_dome}) and (\textit{concept\_athlete-\_joey\_harrington}, \textit{concept:athleteplaysforteam}, \textit{concept\_sportsteam\_falcons}), provide a vivid example that a person or an athlete with \textit{concept\_stadiumoreventvenue\_georgia\_dome} as his or her home stadium might play for the team \textit{concept\_sportsteam\_falcons}. We have such examples more than one, like \textit{concept\_athlete\_roddy\_white}'s and \textit{concept\_athlete\_quarterback\_matt\_ryan}'s. The entity \textit{concept\_sportsleague\_nfl} cannot help us differentiate the true answer from other NFL teams, but it can at least exclude those non-NFL teams. In a word, our subgraph-structured representation can well capture the relational and compositional reasoning pattern.

%% file: p5_conclusion.tex
\section{Conclusion}

We introduce an attentive message passing mechanism on graphs under the notion of attentive awareness, inspired by the phenomenon of consciousness, to model the iterative compositional reasoning pattern by forming a compact query-dependent subgraph. We propose an attentive computation framework with three flow-based layer to combine GNNs' representation power with explicit reasoning process, and further reduce the complexity when applying GNNs to large-scale graphs. It is worth mentioning that our framework is not limited to knowledge graph reasoning, but has a wider applicability to large-scale graph-based computation with a few input-dependent nodes and edges involved each time.

%% file: appendix.tex
\section{Appendix}

\setlength{\tabcolsep}{3.5pt}

\subsection{Hyperparameter settings}

\begin{table}[h]
  \caption{The standard hyperparameter settings we use for each dataset plus their training time for one epoch. For the experimental analysis, we only adjust one hyperparameter and keep the remaining fixed at the standard setting. For NELL995, the training time per epoch means the average time cost of the 12 single-query-relation tasks.}
  \label{tab:hyperparameter_settings}
  \centering
  \begin{tabular}{l|cccccc}
    \noalign{\hrule height 1.5pt}
    Hyperparameter & FB15K-237 & FB15K & WN18RR & WN18 & YAGO3-10 & NELL995  \\
    \hline
    \textit{batch\_size} & 80 & 80 & 100 & 100 & 100 & 10 \\
    \textit{n\_dims\_att} & 50 & 50 & 50 & 50 & 50 & 200 \\
    \textit{n\_dims} & 100 & 100 & 100 & 100 & 100 & 200 \\
    \hline
    \textit{max\_sampled\_edges\_per\_step} & 10000 & 10000 & 10000 & 10000 & 10000 & 10000 \\
    \hline
    \textit{max\_attended\_nodes\_per\_step} & 20 & 20 & 20 & 20 & 20 & 100 \\
    \textit{max\_sampled\_edges\_per\_node} & 200 & 200 & 200 & 200 & 200 & 1000 \\
    \textit{max\_seen\_nodes\_per\_step} & 200 & 200 & 200 & 200 & 200 & 1000 \\
    \hline
    \textit{n\_steps\_of\_u\_flow} & 2 & 1 & 2 & 1 & 1 & 1 \\
    \textit{n\_steps\_of\_c\_flow} & 6 & 6 & 8 & 8 & 6 & 5 \\
    \hline
    \textit{learning\_rate} & 0.001 & 0.001 & 0.001 & 0.001 & 0.0001 & 0.001 \\
    \textit{optimizer} & Adam & Adam & Adam & Adam & Adam & Adam \\ 
    \textit{grad\_clipnorm} & 1 & 1 & 1 & 1 & 1 & 1 \\
    \textit{n\_epochs} & 1 & 1 & 1 & 1 & 1 & 3 \\
    \hline
    \hline
    Training time per epoch (h) & 25.7 & 63.7 & 4.3 & 8.5 & 185.0 & 0.12 \\
    \noalign{\hrule height 1.5pt}
  \end{tabular}
\end{table}

Our hyperparameters can be categorized into three groups: 
\begin{itemize}
    \item The normal hyperparameters, including \textit{batch\_size}, \textit{n\_dims\_att}, \textit{n\_dims}, \textit{learning\_rate}, \textit{grad\_clipnorm}, and \textit{n\_epochs}. Here, we set a smaller dimension, \textit{n\_dims\_att}, for the attention flow computation, as it uses more edges for computation than the message passing uses in the consciousness flow layer, and also intuitively, it does not need to propagate high-dimensional messages but only compute a scalar score for each of the sampled neighbor nodes, in concert with the idea in the key-value mechanism \cite{Bengio2017TheCP}. We set $n\_epochs = 1$ in most cases, indicating that our model needs to be trained only for one epoch due to its fast convergence.
    \item The hyperparameters that are in charge of controlling the sampling-attending horizon, including \textit{max\_sampled\_edges\_per\_step} that controls the maximum number to sample edges per step per query for the message passing in the unconsciousness flow layer, and \textit{max\_sampled\_edges\_per\_node}, \textit{max\_attended\_nodes\_per\_step} and \textit{max\_seen\_nodes\_per\_step} that control the maximum number to sample edges connected to each current node per step per query, the maximum number of current nodes to attend from per step per query, and the maximum number of neighbor nodes to attend to per step per query in the consciousness flow layer.
    \item The hyperparameters that are in charge of controlling the searching horizon, including \textit{n\_steps\_of\_u\_flow} representing the number of steps to run the unconcsiousness flow, and \textit{n\_steps\_of\_c\_flow} representing the number of steps to run the consciousness flow.
\end{itemize}
Note that we choose these hyperparameters not only by their performances but also the computation resources available to us. In some cases, to deal with a very large knowledge graph with limited resources, we need to make a trade-off between the efficiency and the effectiveness. For example, each of NELL995's single-query-relation tasks has a small training set, though still with a large graph, so we can reduce the batch size in favor of affording larger dimensions and a larger sampling-attending horizon without any concern for waiting too long to finish one epoch. 

\subsection{Other experimental analysis}

\begin{figure}[h]
  \centering
  \includegraphics[width=\textwidth]{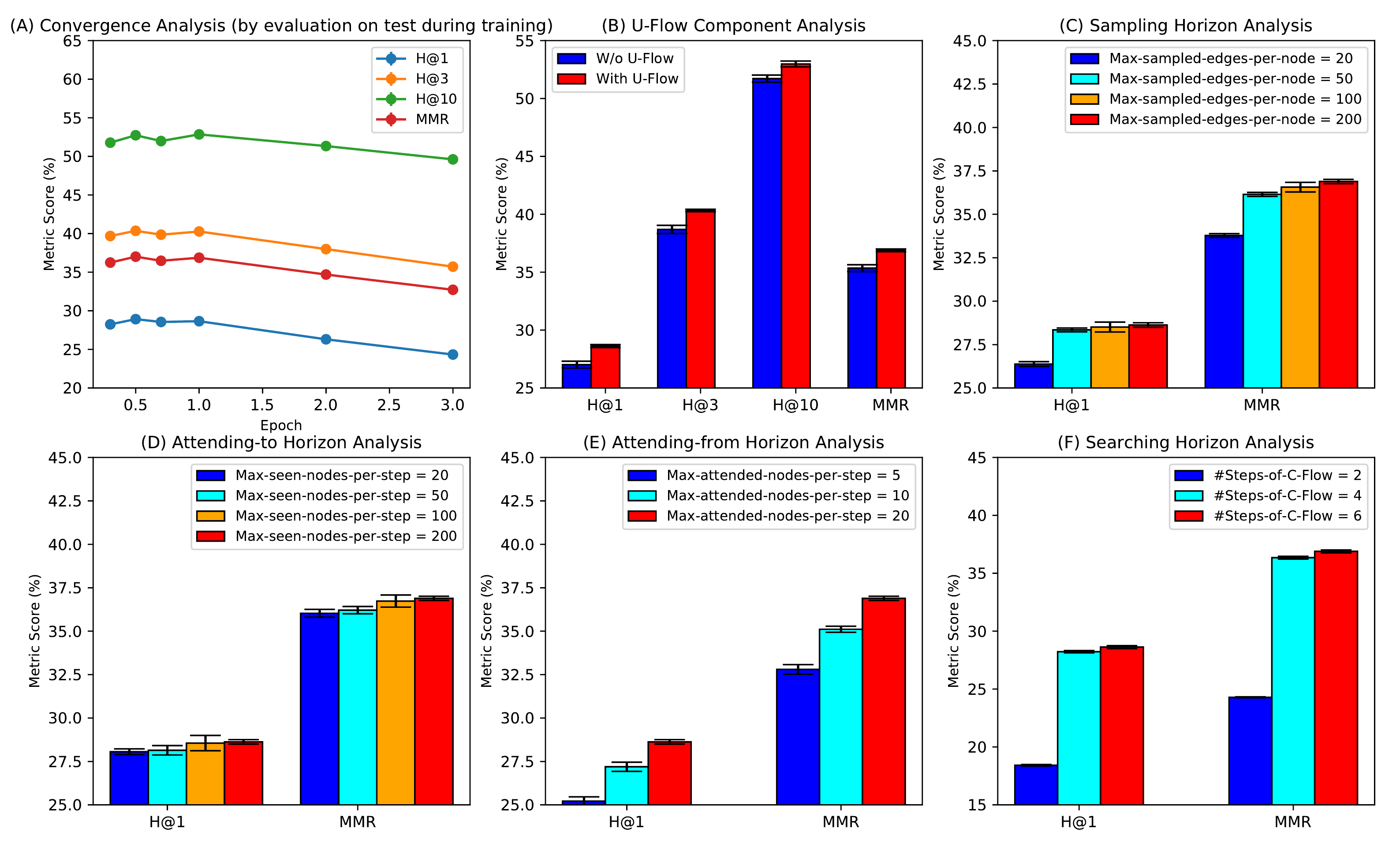}
  \caption{Experimental analysis on FB15K-237: (A) During training we pick six model snapshots at time points of 0.3, 0.5, 0.7, 1, 2, and 3 epochs and evaluate them on test; (B) The \textit{w/o U-Flow} uses zero step to run U-Flow, while the \textit{with U-Flow} uses two steps; (C)-(F) are for the sampling, attending and searching horizon analysis based on the standard hyperparameter settings listed in the appendix.}
  \label{fig:fb237_analysis}
\end{figure}

\begin{figure}[h]
  \centering
  \includegraphics[width=\textwidth]{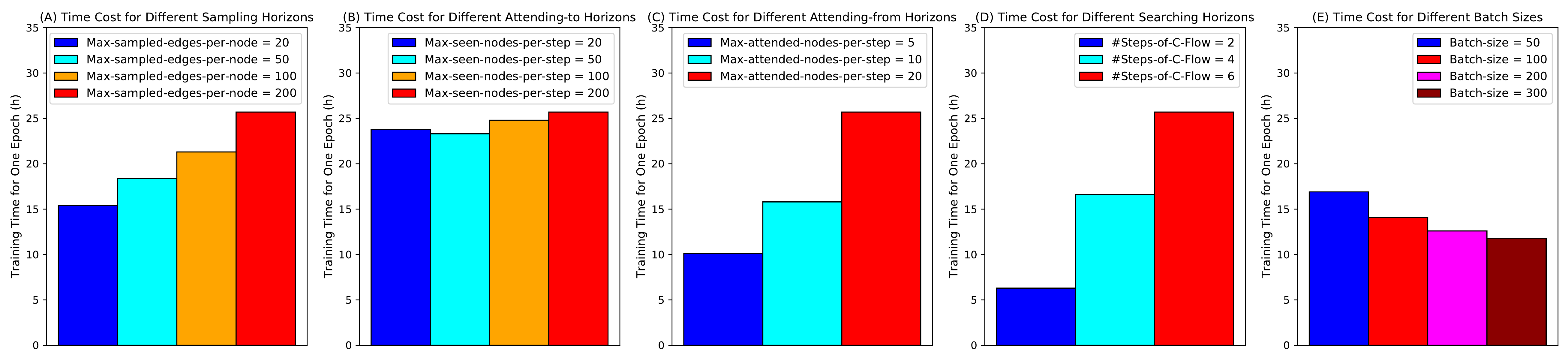}
  \caption{Analysis of time cost on FB15K-237: (A)-(D) measure the training time for one epoch on different horizon settings corresponding to Figure \ref{fig:fb237_analysis}(C)-(F); (E) measures the training time for one epoch for different batch sizes using the same horizon setting, which is \textit{Max-sampled-edges-per-node}=20, \textit{Max-seen-nodes-per-step}=20, \textit{Max-attended-nodes-per-step}=20, and \textit{\#Steps-of-C-Flow}=6.}
  \label{fig:time_cost_fb237}
\end{figure}

See Figure \ref{fig:fb237_analysis},\ref{fig:time_cost_fb237}.

\subsection{Other visualization}

\textbf{For the AthletePlaysInLeague task}

\begin{figure}[h]
  \hspace{-35pt}
  \includegraphics[width=1.2\textwidth]{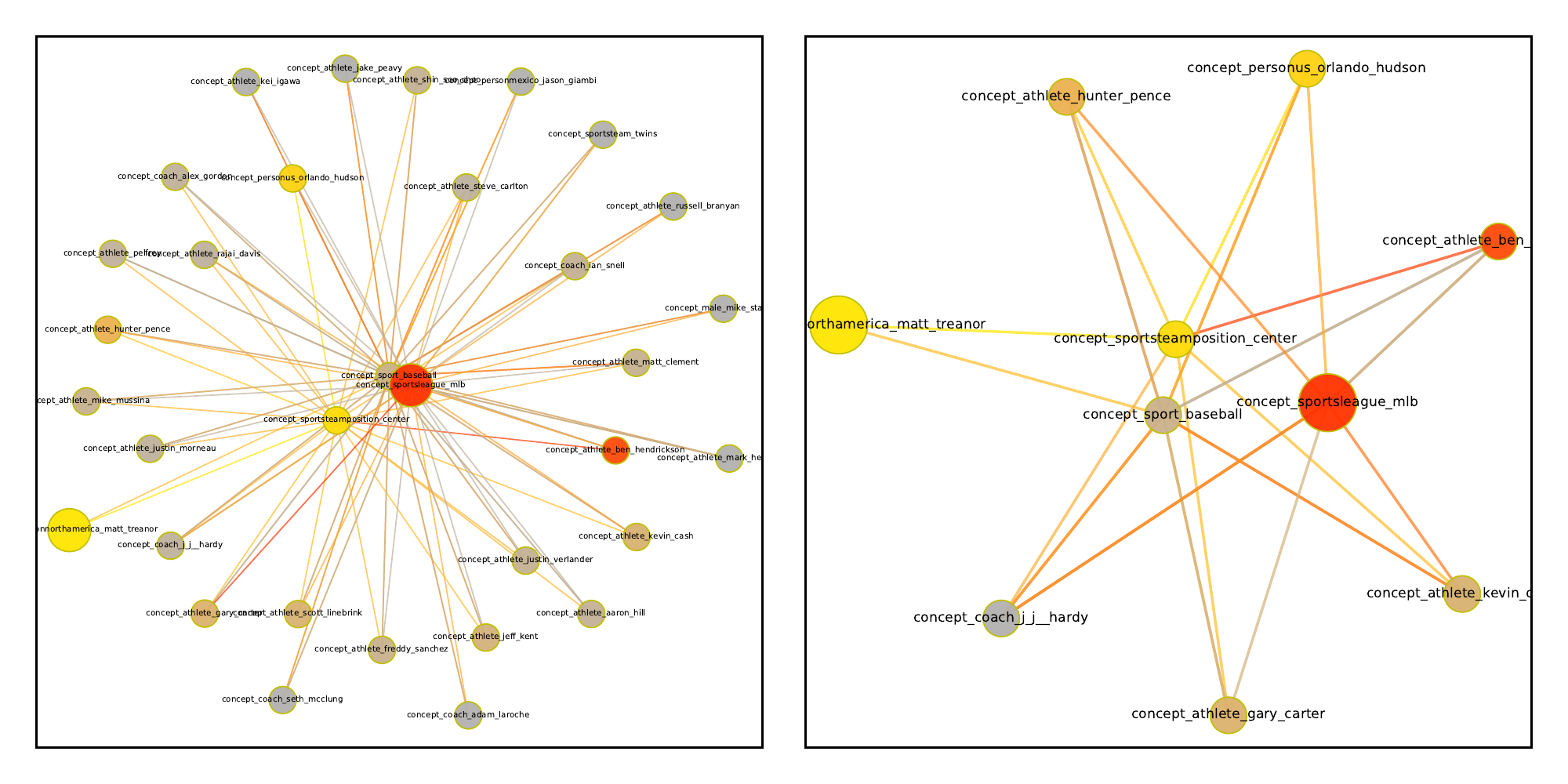}
  \caption{\textbf{AthletePlaysInLeague}. The head is \textit{}, the query relation is \textit{concept:athleteplaysinleague}, and the desired tail is \textit{}. The left is a full subgraph derived with $max\_attended\_nodes\_per\_step = 20$, and the right is a further extracted subgraph from the left based on attention. The big yellow node represents the head, and the big red node represents the tail. Colors indicate how important a node is attended to in a local subgraph. Grey means less important, yellow means it is more attended during the early steps, and red means it is more attended when getting close to the final step.}
  \label{fig:athleteplaysinleague}
\end{figure}

\begin{lstlisting}[basicstyle=\sffamily\scriptsize]
Query: (concept_personnorthamerica_matt_treanor, concept:athleteplaysinleague, concept_sportsleague_mlb)

Selected key edges:
concept_personnorthamerica_matt_treanor, concept:athleteflyouttosportsteamposition, concept_sportsteamposition_center
concept_personnorthamerica_matt_treanor, concept:athleteplayssport, concept_sport_baseball
concept_sportsteamposition_center, concept:athleteflyouttosportsteamposition_inv, concept_personus_orlando_hudson
concept_sportsteamposition_center, concept:athleteflyouttosportsteamposition_inv, concept_athlete_ben_hendrickson
concept_sportsteamposition_center, concept:athleteflyouttosportsteamposition_inv, concept_coach_j_j__hardy
concept_sportsteamposition_center, concept:athleteflyouttosportsteamposition_inv, concept_athlete_hunter_pence
concept_sport_baseball, concept:athleteplayssport_inv, concept_personus_orlando_hudson
concept_sport_baseball, concept:athleteplayssport_inv, concept_athlete_ben_hendrickson
concept_sport_baseball, concept:athleteplayssport_inv, concept_coach_j_j__hardy
concept_sport_baseball, concept:athleteplayssport_inv, concept_athlete_hunter_pence
concept_personus_orlando_hudson, concept:athleteplaysinleague, concept_sportsleague_mlb
concept_personus_orlando_hudson, concept:athleteplayssport, concept_sport_baseball
concept_athlete_ben_hendrickson, concept:coachesinleague, concept_sportsleague_mlb
concept_athlete_ben_hendrickson, concept:athleteplayssport, concept_sport_baseball
concept_coach_j_j__hardy, concept:coachesinleague, concept_sportsleague_mlb
concept_coach_j_j__hardy, concept:athleteplaysinleague, concept_sportsleague_mlb
concept_coach_j_j__hardy, concept:athleteplayssport, concept_sport_baseball
concept_athlete_hunter_pence, concept:athleteplaysinleague, concept_sportsleague_mlb
concept_athlete_hunter_pence, concept:athleteplayssport, concept_sport_baseball
concept_sportsleague_mlb, concept:coachesinleague_inv, concept_athlete_ben_hendrickson
concept_sportsleague_mlb, concept:coachesinleague_inv, concept_coach_j_j__hardy
\end{lstlisting}

\textbf{For the AthleteHomeStadium task}

\begin{figure}[h]
  \hspace{-35pt}
  \includegraphics[width=1.2\textwidth]{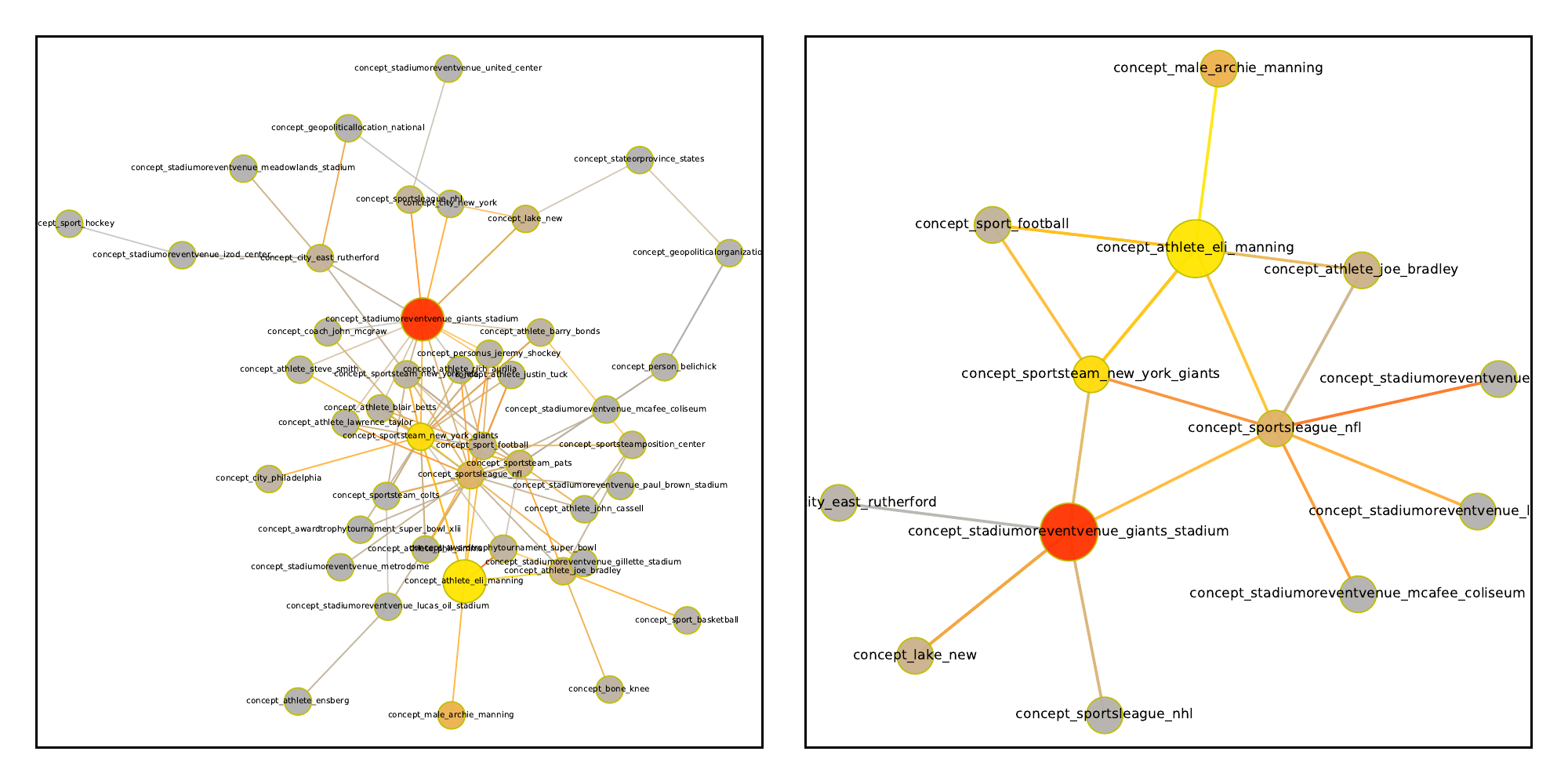}
  \caption{\textbf{AthleteHomeStadium}. The head is \textit{concept\_athlete\_eli\_manning}, the query relation is \textit{concept:athletehomestadium}, and the desired tail is \textit{concept\_stadiumoreventvenue\_giants\_stadium}. The left is a full subgraph derived with $max\_attended\_nodes\_per\_step = 20$, and the right is a further extracted subgraph from the left based on attention. The big yellow node represents the head, and the big red node represents the tail. Colors indicate how important a node is attended to in a local subgraph. Grey means less important, yellow means it is more attended during the early steps, and red means it is more attended when getting close to the final step.}
  \label{fig:}
\end{figure}

\begin{lstlisting}[basicstyle=\sffamily\scriptsize]
Query: (concept_athlete_eli_manning, concept:athletehomestadium, concept_stadiumoreventvenue_giants_stadium)

Selected key edges:
concept_athlete_eli_manning, concept:personbelongstoorganization, concept_sportsteam_new_york_giants
concept_athlete_eli_manning, concept:athleteplaysforteam, concept_sportsteam_new_york_giants
concept_athlete_eli_manning, concept:athleteledsportsteam, concept_sportsteam_new_york_giants
concept_athlete_eli_manning, concept:athleteplaysinleague, concept_sportsleague_nfl
concept_athlete_eli_manning, concept:fatherofperson_inv, concept_male_archie_manning
concept_sportsteam_new_york_giants, concept:teamplaysinleague, concept_sportsleague_nfl
concept_sportsteam_new_york_giants, concept:teamhomestadium, concept_stadiumoreventvenue_giants_stadium
concept_sportsteam_new_york_giants, concept:personbelongstoorganization_inv, concept_athlete_eli_manning
concept_sportsteam_new_york_giants, concept:athleteplaysforteam_inv, concept_athlete_eli_manning
concept_sportsteam_new_york_giants, concept:athleteledsportsteam_inv, concept_athlete_eli_manning
concept_sportsleague_nfl, concept:teamplaysinleague_inv, concept_sportsteam_new_york_giants
concept_sportsleague_nfl, concept:agentcompeteswithagent, concept_sportsleague_nfl
concept_sportsleague_nfl, concept:agentcompeteswithagent_inv, concept_sportsleague_nfl
concept_sportsleague_nfl, concept:leaguestadiums, concept_stadiumoreventvenue_giants_stadium
concept_sportsleague_nfl, concept:athleteplaysinleague_inv, concept_athlete_eli_manning
concept_male_archie_manning, concept:fatherofperson, concept_athlete_eli_manning
concept_sportsleague_nfl, concept:leaguestadiums, concept_stadiumoreventvenue_paul_brown_stadium
concept_stadiumoreventvenue_giants_stadium, concept:teamhomestadium_inv, concept_sportsteam_new_york_giants
concept_stadiumoreventvenue_giants_stadium, concept:leaguestadiums_inv, concept_sportsleague_nfl
concept_stadiumoreventvenue_giants_stadium, concept:proxyfor_inv, concept_city_east_rutherford
concept_city_east_rutherford, concept:proxyfor, concept_stadiumoreventvenue_giants_stadium
concept_stadiumoreventvenue_paul_brown_stadium, concept:leaguestadiums_inv, concept_sportsleague_nfl
\end{lstlisting}

\textbf{For the AthletePlaysSport task}

\begin{figure}[h]
  \hspace{-35pt}
  \includegraphics[width=1.2\textwidth]{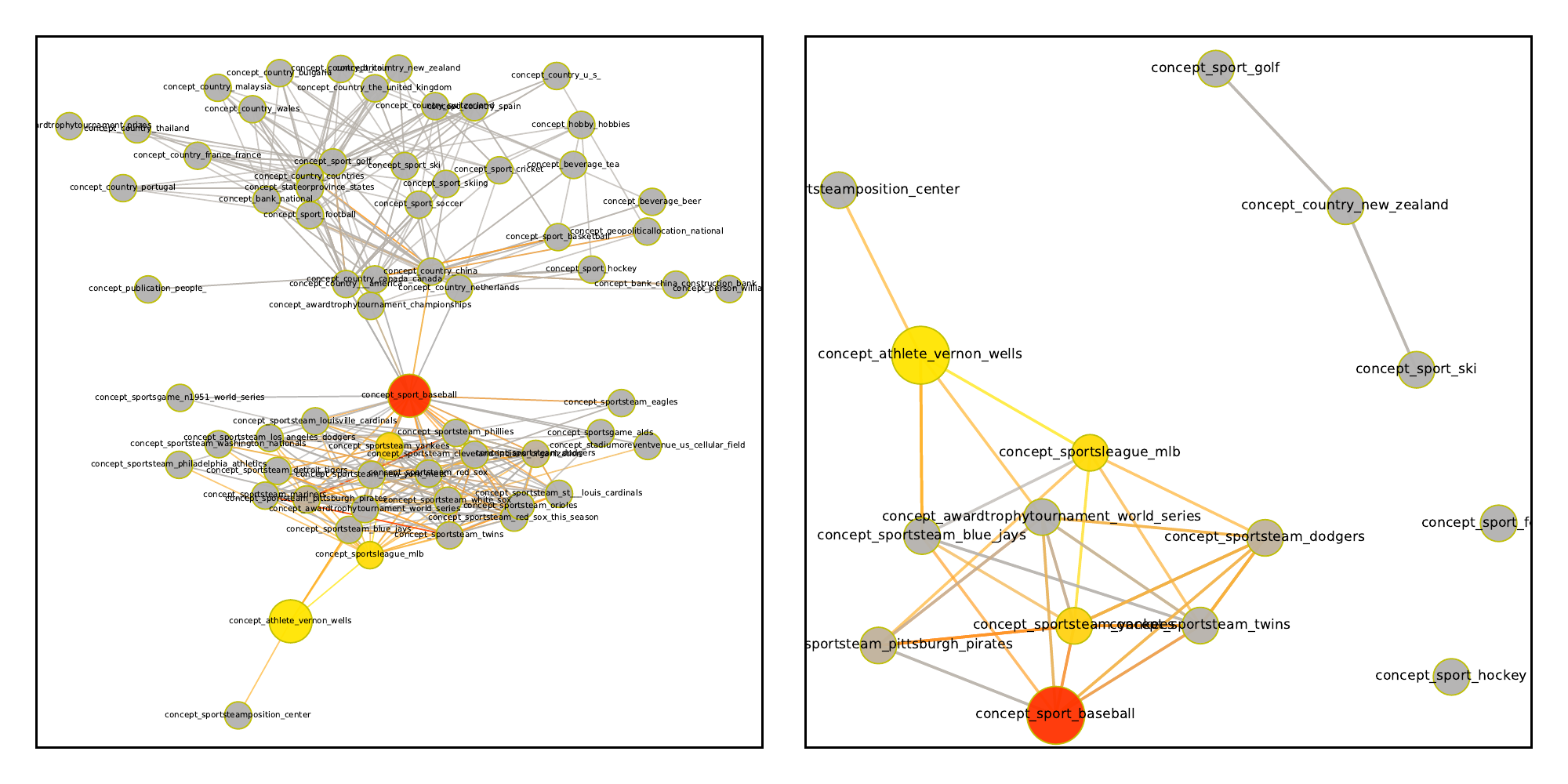}
  \caption{\textbf{AthletePlaysSport}. The head is \textit{concept\_athlete\_vernon\_wells}, the query relation is \textit{concept:athleteplayssport}, and the desired tail is \textit{concept\_sport\_baseball}. The left is a full subgraph derived with $max\_attended\_nodes\_per\_step = 20$, and the right is a further extracted subgraph from the left based on attention. The big yellow node represents the head, and the big red node represents the tail. Colors indicate how important a node is attended to in a local subgraph. Grey means less important, yellow means it is more attended during the early steps, and red means it is more attended when getting close to the final step.}
  \label{fig:}
\end{figure}

\begin{lstlisting}[basicstyle=\sffamily\scriptsize]
Query: (concept_athlete_vernon_wells, concept:athleteplayssport, concept_sport_baseball)

Selected key edges:
concept_athlete_vernon_wells, concept:athleteplaysinleague, concept_sportsleague_mlb
concept_athlete_vernon_wells, concept:coachwontrophy, concept_awardtrophytournament_world_series
concept_athlete_vernon_wells, concept:agentcollaborateswithagent_inv, concept_sportsteam_blue_jays
concept_athlete_vernon_wells, concept:personbelongstoorganization, concept_sportsteam_blue_jays
concept_athlete_vernon_wells, concept:athleteplaysforteam, concept_sportsteam_blue_jays
concept_athlete_vernon_wells, concept:athleteledsportsteam, concept_sportsteam_blue_jays
concept_sportsleague_mlb, concept:teamplaysinleague_inv, concept_sportsteam_dodgers
concept_sportsleague_mlb, concept:teamplaysinleague_inv, concept_sportsteam_yankees
concept_sportsleague_mlb, concept:teamplaysinleague_inv, concept_sportsteam_pittsburgh_pirates
concept_awardtrophytournament_world_series, concept:teamwontrophy_inv, concept_sportsteam_dodgers
concept_awardtrophytournament_world_series, concept:teamwontrophy_inv, concept_sportsteam_yankees
concept_awardtrophytournament_world_series, concept:awardtrophytournamentisthechampionshipgameofthenationalsport,
    concept_sport_baseball
concept_awardtrophytournament_world_series, concept:teamwontrophy_inv, concept_sportsteam_pittsburgh_pirates
concept_sportsteam_blue_jays, concept:teamplaysinleague, concept_sportsleague_mlb
concept_sportsteam_blue_jays, concept:teamplaysagainstteam, concept_sportsteam_yankees
concept_sportsteam_blue_jays, concept:teamplayssport, concept_sport_baseball
concept_sportsteam_dodgers, concept:teamplaysagainstteam, concept_sportsteam_yankees
concept_sportsteam_dodgers, concept:teamplaysagainstteam_inv, concept_sportsteam_yankees
concept_sportsteam_dodgers, concept:teamwontrophy, concept_awardtrophytournament_world_series
concept_sportsteam_dodgers, concept:teamplayssport, concept_sport_baseball
concept_sportsteam_yankees, concept:teamplaysagainstteam, concept_sportsteam_dodgers
concept_sportsteam_yankees, concept:teamplaysagainstteam_inv, concept_sportsteam_dodgers
concept_sportsteam_yankees, concept:teamwontrophy, concept_awardtrophytournament_world_series
concept_sportsteam_yankees, concept:teamplayssport, concept_sport_baseball
concept_sportsteam_yankees, concept:teamplaysagainstteam, concept_sportsteam_pittsburgh_pirates
concept_sportsteam_yankees, concept:teamplaysagainstteam_inv, concept_sportsteam_pittsburgh_pirates
concept_sport_baseball, concept:teamplayssport_inv, concept_sportsteam_dodgers
concept_sport_baseball, concept:teamplayssport_inv, concept_sportsteam_yankees
concept_sport_baseball, concept:awardtrophytournamentisthechampionshipgameofthenationalsport_inv, 
    concept_awardtrophytournament_world_series
concept_sport_baseball, concept:teamplayssport_inv, concept_sportsteam_pittsburgh_pirates
concept_sportsteam_pittsburgh_pirates, concept:teamplaysagainstteam, concept_sportsteam_yankees
concept_sportsteam_pittsburgh_pirates, concept:teamplaysagainstteam_inv, concept_sportsteam_yankees
concept_sportsteam_pittsburgh_pirates, concept:teamwontrophy, concept_awardtrophytournament_world_series
concept_sportsteam_pittsburgh_pirates, concept:teamplayssport, concept_sport_baseball
\end{lstlisting}

\textbf{For the TeamPlaysSport task}

\begin{figure}[h]
  \hspace{-35pt}
  \includegraphics[width=1.2\textwidth]{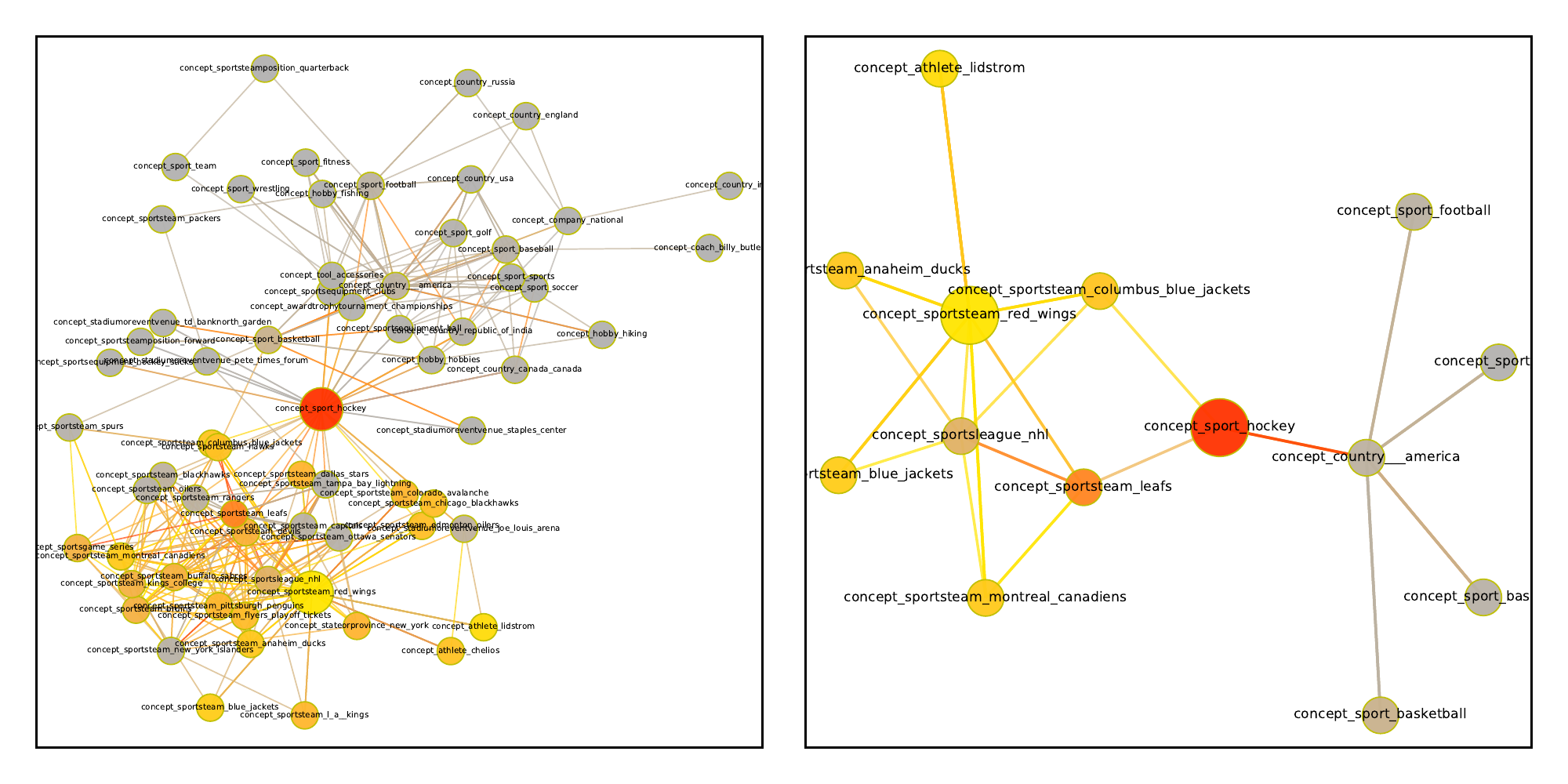}
  \caption{\textbf{TeamPlaysSport}. The head is \textit{concept\_sportsteam\_red\_wings}, the query relation is \textit{concept:teamplayssport}, and the desired tail is \textit{concept\_sport\_hockey}. The left is a full subgraph derived with $max\_attended\_nodes\_per\_step = 20$, and the right is a further extracted subgraph from the left based on attention. The big yellow node represents the head, and the big red node represents the tail. Colors indicate how important a node is attended to in a local subgraph. Grey means less important, yellow means it is more attended during the early steps, and red means it is more attended when getting close to the final step.}
  \label{fig:}
\end{figure}

\begin{lstlisting}[basicstyle=\sffamily\scriptsize]
Query: (concept_sportsteam_red_wings, concept:teamplayssport, concept_sport_hockey)

Selected key edges:
concept_sportsteam_red_wings, concept:teamplaysagainstteam, concept_sportsteam_montreal_canadiens
concept_sportsteam_red_wings, concept:teamplaysagainstteam_inv, concept_sportsteam_montreal_canadiens
concept_sportsteam_red_wings, concept:teamplaysagainstteam, concept_sportsteam_blue_jackets
concept_sportsteam_red_wings, concept:teamplaysagainstteam_inv, concept_sportsteam_blue_jackets
concept_sportsteam_red_wings, concept:worksfor_inv, concept_athlete_lidstrom
concept_sportsteam_red_wings, concept:organizationhiredperson, concept_athlete_lidstrom
concept_sportsteam_red_wings, concept:athleteplaysforteam_inv, concept_athlete_lidstrom
concept_sportsteam_red_wings, concept:athleteledsportsteam_inv, concept_athlete_lidstrom
concept_sportsteam_montreal_canadiens, concept:teamplaysagainstteam, concept_sportsteam_red_wings
concept_sportsteam_montreal_canadiens, concept:teamplaysagainstteam_inv, concept_sportsteam_red_wings
concept_sportsteam_montreal_canadiens, concept:teamplaysinleague, concept_sportsleague_nhl
concept_sportsteam_montreal_canadiens, concept:teamplaysagainstteam, concept_sportsteam_leafs
concept_sportsteam_montreal_canadiens, concept:teamplaysagainstteam_inv, concept_sportsteam_leafs
concept_sportsteam_blue_jackets, concept:teamplaysagainstteam, concept_sportsteam_red_wings
concept_sportsteam_blue_jackets, concept:teamplaysagainstteam_inv, concept_sportsteam_red_wings
concept_sportsteam_blue_jackets, concept:teamplaysinleague, concept_sportsleague_nhl
concept_athlete_lidstrom, concept:worksfor, concept_sportsteam_red_wings
concept_athlete_lidstrom, concept:organizationhiredperson_inv, concept_sportsteam_red_wings
concept_athlete_lidstrom, concept:athleteplaysforteam, concept_sportsteam_red_wings
concept_athlete_lidstrom, concept:athleteledsportsteam, concept_sportsteam_red_wings
concept_sportsteam_red_wings, concept:teamplaysinleague, concept_sportsleague_nhl
concept_sportsteam_red_wings, concept:teamplaysagainstteam, concept_sportsteam_leafs
concept_sportsteam_red_wings, concept:teamplaysagainstteam_inv, concept_sportsteam_leafs
concept_sportsleague_nhl, concept:agentcompeteswithagent, concept_sportsleague_nhl
concept_sportsleague_nhl, concept:agentcompeteswithagent_inv, concept_sportsleague_nhl
concept_sportsleague_nhl, concept:teamplaysinleague_inv, concept_sportsteam_leafs
concept_sportsteam_leafs, concept:teamplaysinleague, concept_sportsleague_nhl
concept_sportsteam_leafs, concept:teamplayssport, concept_sport_hockey

\end{lstlisting}

\textbf{For the OrganizationHeadQuarteredInCity task}

\begin{figure}[h]
  \hspace{-35pt}
  \includegraphics[width=1.2\textwidth]{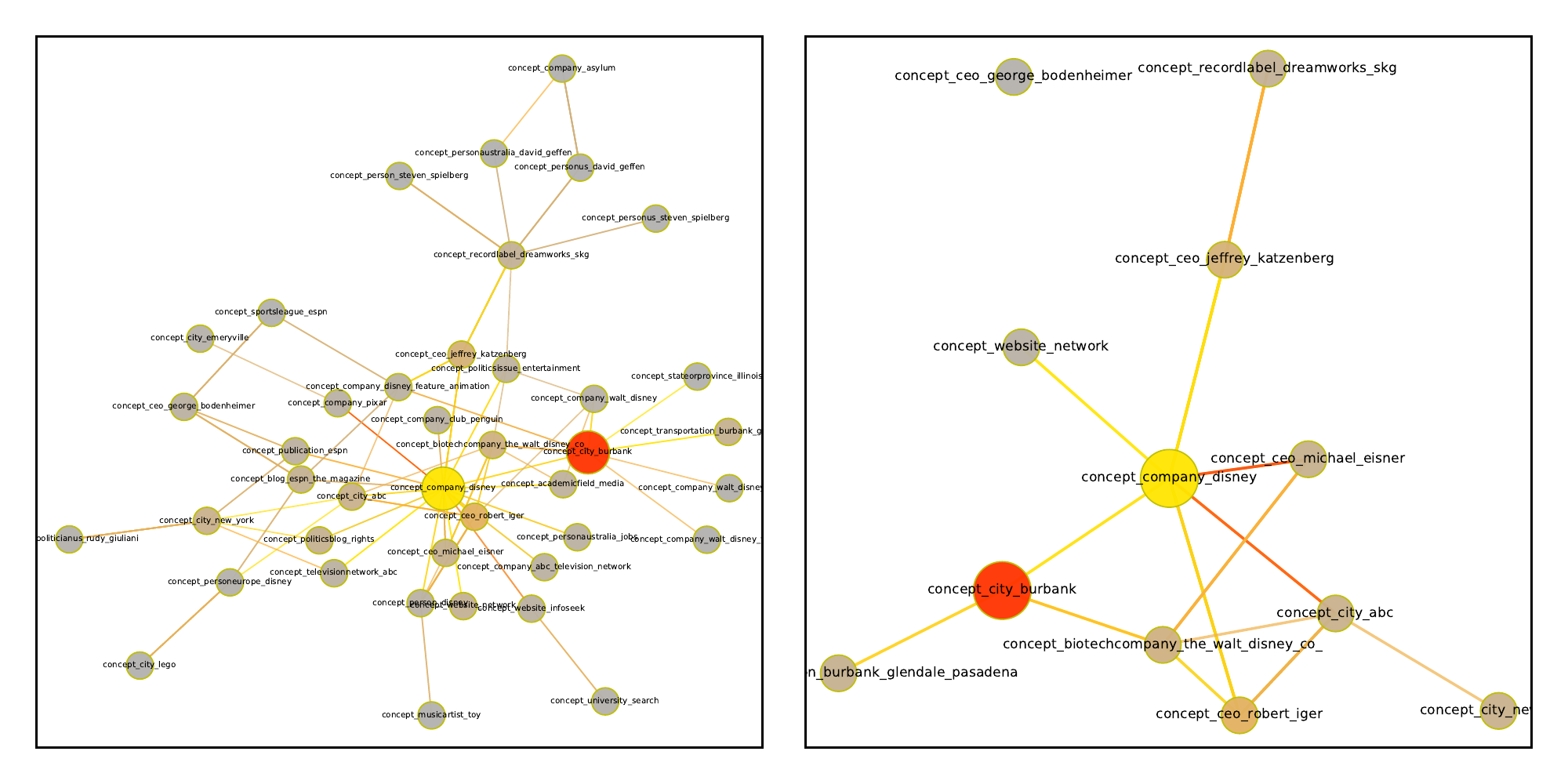}
  \caption{\textbf{OrganizationHeadQuarteredInCity}. The head is \textit{concept\_company\_disney}, the query relation is \textit{concept:organizationheadquarteredincity}, and the desired tail is \textit{concept\_city\_burbank}. The left is a full subgraph derived with $max\_attended\_nodes\_per\_step = 20$, and the right is a further extracted subgraph from the left based on attention. The big yellow node represents the head, and the big red node represents the tail. Colors indicate how important a node is attended to in a local subgraph. Grey means less important, yellow means it is more attended during the early steps, and red means it is more attended when getting close to the final step.}
  \label{fig:}
\end{figure}

\begin{lstlisting}[basicstyle=\sffamily\scriptsize]
Query: (concept_company_disney, concept:organizationheadquarteredincity, concept_city_burbank)

Selected key edges:
concept_company_disney, concept:headquarteredin, concept_city_burbank
concept_company_disney, concept:subpartoforganization_inv, concept_website_network
concept_company_disney, concept:worksfor_inv, concept_ceo_robert_iger
concept_company_disney, concept:proxyfor_inv, concept_ceo_robert_iger
concept_company_disney, concept:personleadsorganization_inv, concept_ceo_robert_iger
concept_company_disney, concept:ceoof_inv, concept_ceo_robert_iger
concept_company_disney, concept:personleadsorganization_inv, concept_ceo_jeffrey_katzenberg
concept_company_disney, concept:organizationhiredperson, concept_ceo_jeffrey_katzenberg
concept_company_disney, concept:organizationterminatedperson, concept_ceo_jeffrey_katzenberg
concept_city_burbank, concept:headquarteredin_inv, concept_company_disney
concept_city_burbank, concept:headquarteredin_inv, concept_biotechcompany_the_walt_disney_co_
concept_website_network, concept:subpartoforganization, concept_company_disney
concept_ceo_robert_iger, concept:worksfor, concept_company_disney
concept_ceo_robert_iger, concept:proxyfor, concept_company_disney
concept_ceo_robert_iger, concept:personleadsorganization, concept_company_disney
concept_ceo_robert_iger, concept:ceoof, concept_company_disney
concept_ceo_robert_iger, concept:topmemberoforganization, concept_biotechcompany_the_walt_disney_co_
concept_ceo_robert_iger, concept:organizationterminatedperson_inv, concept_biotechcompany_the_walt_disney_co_
concept_ceo_jeffrey_katzenberg, concept:personleadsorganization, concept_company_disney
concept_ceo_jeffrey_katzenberg, concept:organizationhiredperson_inv, concept_company_disney
concept_ceo_jeffrey_katzenberg, concept:organizationterminatedperson_inv, concept_company_disney
concept_ceo_jeffrey_katzenberg, concept:worksfor, concept_recordlabel_dreamworks_skg
concept_ceo_jeffrey_katzenberg, concept:topmemberoforganization, concept_recordlabel_dreamworks_skg
concept_ceo_jeffrey_katzenberg, concept:organizationterminatedperson_inv, concept_recordlabel_dreamworks_skg
concept_ceo_jeffrey_katzenberg, concept:ceoof, concept_recordlabel_dreamworks_skg
concept_biotechcompany_the_walt_disney_co_, concept:headquarteredin, concept_city_burbank
concept_biotechcompany_the_walt_disney_co_, concept:organizationheadquarteredincity, concept_city_burbank
concept_recordlabel_dreamworks_skg, concept:worksfor_inv, concept_ceo_jeffrey_katzenberg
concept_recordlabel_dreamworks_skg, concept:topmemberoforganization_inv, concept_ceo_jeffrey_katzenberg
concept_recordlabel_dreamworks_skg, concept:organizationterminatedperson, concept_ceo_jeffrey_katzenberg
concept_recordlabel_dreamworks_skg, concept:ceoof_inv, concept_ceo_jeffrey_katzenberg
concept_city_burbank, concept:airportincity_inv, concept_transportation_burbank_glendale_pasadena
concept_transportation_burbank_glendale_pasadena, concept:airportincity, concept_city_burbank
\end{lstlisting}

\textbf{For the WorksFor task}

\begin{figure}[h]
  \hspace{-35pt}
  \includegraphics[width=1.2\textwidth]{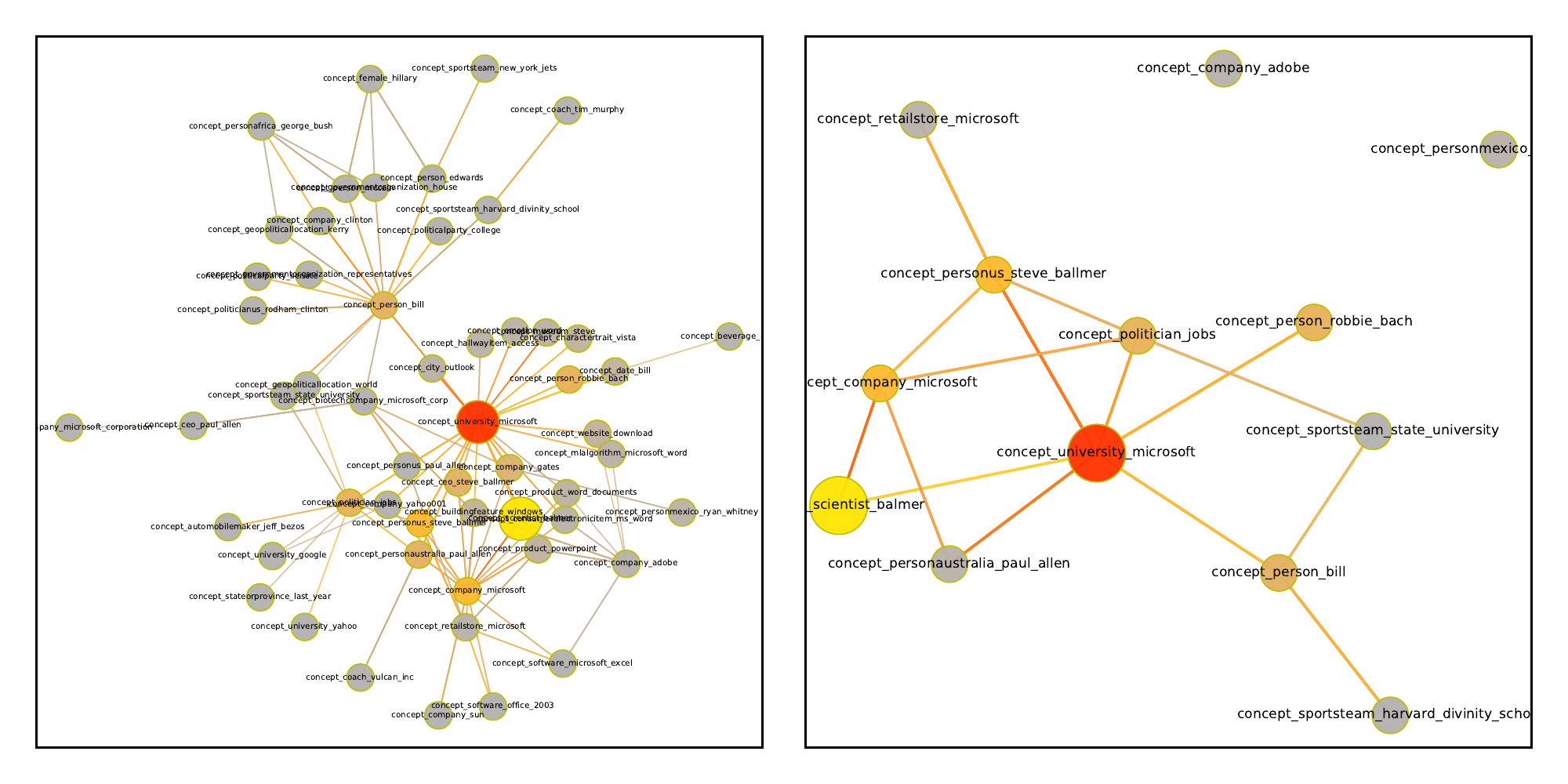}
  \caption{\textbf{WorksFor}. The head is \textit{concept\_scientist\_balmer}, the query relation is \textit{concept:worksfor}, and the desired tail is \textit{concept\_university\_microsoft}. The left is a full subgraph derived with $max\_attended\_nodes\_per\_step = 20$, and the right is a further extracted subgraph from the left based on attention. The big yellow node represents the head, and the big red node represents the tail. Colors indicate how important a node is attended to in a local subgraph. Grey means less important, yellow means it is more attended during the early steps, and red means it is more attended when getting close to the final step.}
  \label{fig:}
\end{figure}

\begin{lstlisting}[basicstyle=\sffamily\scriptsize]
Query: (concept_scientist_balmer, concept:worksfor, concept_university_microsoft)

Selected key edges:
concept_scientist_balmer, concept:topmemberoforganization, concept_company_microsoft
concept_scientist_balmer, concept:organizationterminatedperson_inv, concept_university_microsoft
concept_company_microsoft, concept:topmemberoforganization_inv, concept_personus_steve_ballmer
concept_company_microsoft, concept:topmemberoforganization_inv, concept_scientist_balmer
concept_university_microsoft, concept:agentcollaborateswithagent, concept_personus_steve_ballmer
concept_university_microsoft, concept:personleadsorganization_inv, concept_personus_steve_ballmer
concept_university_microsoft, concept:personleadsorganization_inv, concept_person_bill
concept_university_microsoft, concept:organizationterminatedperson, concept_scientist_balmer
concept_university_microsoft, concept:personleadsorganization_inv, concept_person_robbie_bach
concept_personus_steve_ballmer, concept:topmemberoforganization, concept_company_microsoft
concept_personus_steve_ballmer, concept:agentcollaborateswithagent_inv, concept_university_microsoft
concept_personus_steve_ballmer, concept:personleadsorganization, concept_university_microsoft
concept_personus_steve_ballmer, concept:worksfor, concept_university_microsoft
concept_personus_steve_ballmer, concept:proxyfor, concept_retailstore_microsoft
concept_personus_steve_ballmer, concept:subpartof, concept_retailstore_microsoft
concept_personus_steve_ballmer, concept:agentcontrols, concept_retailstore_microsoft
concept_person_bill, concept:personleadsorganization, concept_university_microsoft
concept_person_bill, concept:worksfor, concept_university_microsoft
concept_person_robbie_bach, concept:personleadsorganization, concept_university_microsoft
concept_person_robbie_bach, concept:worksfor, concept_university_microsoft
concept_retailstore_microsoft, concept:proxyfor_inv, concept_personus_steve_ballmer
concept_retailstore_microsoft, concept:subpartof_inv, concept_personus_steve_ballmer
concept_retailstore_microsoft, concept:agentcontrols_inv, concept_personus_steve_ballmer
\end{lstlisting}

\textbf{For the PersonBornInLocation task}

\begin{figure}[h]
  \hspace{-35pt}
  \includegraphics[width=1.2\textwidth]{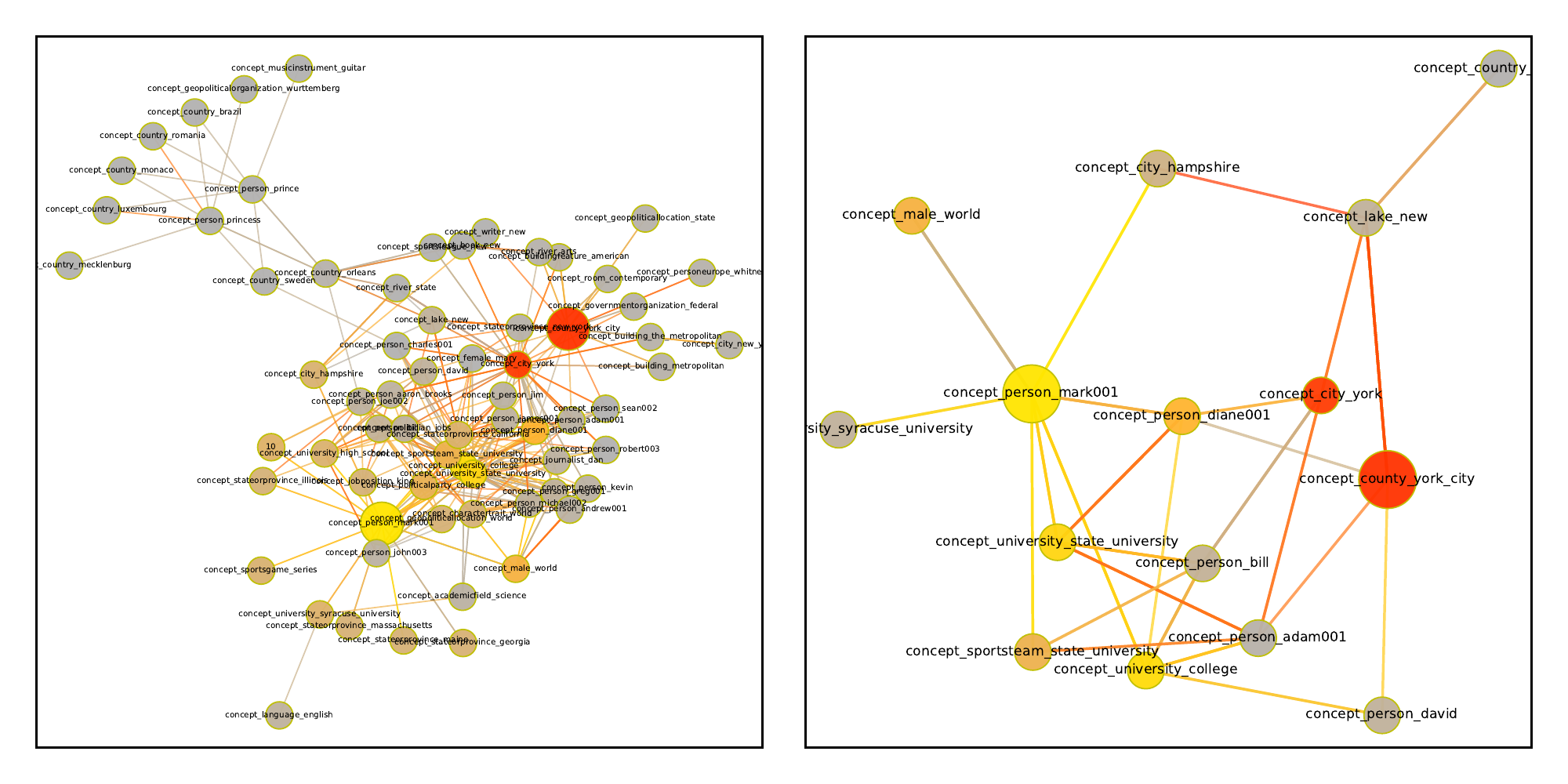}
  \caption{\textbf{PersonBornInLocation}. The head is \textit{concept\_person\_mark001}, the query relation is \textit{concept:personborninlocation}, and the desired tail is \textit{concept\_county\_york\_city}. The left is a full subgraph derived with $max\_attended\_nodes\_per\_step = 20$, and the right is a further extracted subgraph from the left based on attention. The big yellow node represents the head, and the big red node represents the tail. Colors indicate how important a node is attended to in a local subgraph. Grey means less important, yellow means it is more attended during the early steps, and red means it is more attended when getting close to the final step.}
  \label{fig:}
\end{figure}

\begin{lstlisting}[basicstyle=\sffamily\scriptsize]
Query: (concept_person_mark001, concept:personborninlocation, concept_county_york_city)

Selected key edges:
concept_person_mark001, concept:persongraduatedfromuniversity, concept_university_college
concept_person_mark001, concept:persongraduatedschool, concept_university_college
concept_person_mark001, concept:persongraduatedfromuniversity, concept_university_state_university
concept_person_mark001, concept:persongraduatedschool, concept_university_state_university
concept_person_mark001, concept:personbornincity, concept_city_hampshire
concept_person_mark001, concept:hasspouse, concept_person_diane001
concept_person_mark001, concept:hasspouse_inv, concept_person_diane001
concept_university_college, concept:persongraduatedfromuniversity_inv, concept_person_mark001
concept_university_college, concept:persongraduatedschool_inv, concept_person_mark001
concept_university_college, concept:persongraduatedfromuniversity_inv, concept_person_bill
concept_university_college, concept:persongraduatedschool_inv, concept_person_bill
concept_university_state_university, concept:persongraduatedfromuniversity_inv, concept_person_mark001
concept_university_state_university, concept:persongraduatedschool_inv, concept_person_mark001
concept_university_state_university, concept:persongraduatedfromuniversity_inv, concept_person_bill
concept_university_state_university, concept:persongraduatedschool_inv, concept_person_bill
concept_city_hampshire, concept:personbornincity_inv, concept_person_mark001
concept_person_diane001, concept:persongraduatedfromuniversity, concept_university_state_university
concept_person_diane001, concept:persongraduatedschool, concept_university_state_university
concept_person_diane001, concept:hasspouse, concept_person_mark001
concept_person_diane001, concept:hasspouse_inv, concept_person_mark001
concept_person_diane001, concept:personborninlocation, concept_county_york_city
concept_university_state_university, concept:persongraduatedfromuniversity_inv, concept_person_diane001
concept_university_state_university, concept:persongraduatedschool_inv, concept_person_diane001
concept_person_bill, concept:personbornincity, concept_city_york
concept_person_bill, concept:personborninlocation, concept_city_york
concept_person_bill, concept:persongraduatedfromuniversity, concept_university_college
concept_person_bill, concept:persongraduatedschool, concept_university_college
concept_person_bill, concept:persongraduatedfromuniversity, concept_university_state_university
concept_person_bill, concept:persongraduatedschool, concept_university_state_university
concept_city_york, concept:personbornincity_inv, concept_person_bill
concept_city_york, concept:personbornincity_inv, concept_person_diane001
concept_university_college, concept:persongraduatedfromuniversity_inv, concept_person_diane001
concept_person_diane001, concept:personbornincity, concept_city_york
\end{lstlisting}

\textbf{For the PersonLeadsOrganization task}

\begin{figure}[h]
  \hspace{-35pt}
  \includegraphics[width=1.2\textwidth]{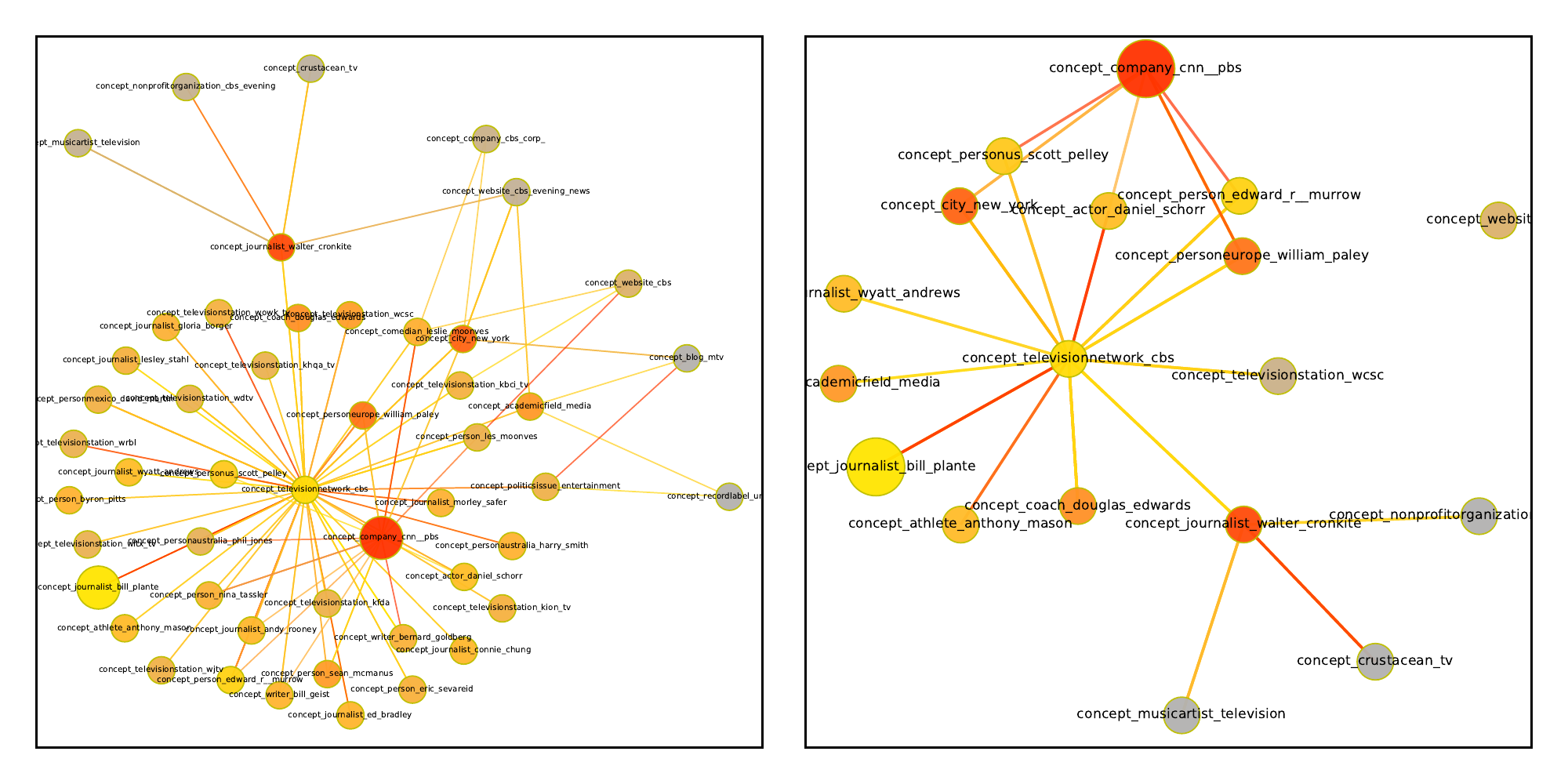}
  \caption{\textbf{PersonLeadsOrganization}. The head is \textit{concept\_journalist\_bill\_plante}, the query relation is \textit{concept:organizationheadquarteredincity}, and the desired tail is \textit{concept\_company\_cnn\_\_pbs}. The left is a full subgraph derived with $max\_attended\_nodes\_per\_step = 20$, and the right is a further extracted subgraph from the left based on attention. The big yellow node represents the head, and the big red node represents the tail. Colors indicate how important a node is attended to in a local subgraph. Grey means less important, yellow means it is more attended during the early steps, and red means it is more attended when getting close to the final step.}
  \label{fig:}
\end{figure}

\begin{lstlisting}[basicstyle=\sffamily\scriptsize]
Query: (concept_journalist_bill_plante, concept:personleadsorganization, concept_company_cnn__pbs)

Selected key edges:
concept_journalist_bill_plante, concept:worksfor, concept_televisionnetwork_cbs
concept_journalist_bill_plante, concept:agentcollaborateswithagent_inv, concept_televisionnetwork_cbs
concept_televisionnetwork_cbs, concept:worksfor_inv, concept_journalist_walter_cronkite
concept_televisionnetwork_cbs, concept:agentcollaborateswithagent, concept_journalist_walter_cronkite
concept_televisionnetwork_cbs, concept:worksfor_inv, concept_personus_scott_pelley
concept_televisionnetwork_cbs, concept:worksfor_inv, concept_actor_daniel_schorr
concept_televisionnetwork_cbs, concept:worksfor_inv, concept_person_edward_r__murrow
concept_televisionnetwork_cbs, concept:agentcollaborateswithagent, concept_person_edward_r__murrow
concept_televisionnetwork_cbs, concept:worksfor_inv, concept_journalist_bill_plante
concept_televisionnetwork_cbs, concept:agentcollaborateswithagent, concept_journalist_bill_plante
concept_journalist_walter_cronkite, concept:worksfor, concept_televisionnetwork_cbs
concept_journalist_walter_cronkite, concept:agentcollaborateswithagent_inv, concept_televisionnetwork_cbs
concept_journalist_walter_cronkite, concept:worksfor, concept_nonprofitorganization_cbs_evening
concept_personus_scott_pelley, concept:worksfor, concept_televisionnetwork_cbs
concept_personus_scott_pelley, concept:personleadsorganization, concept_televisionnetwork_cbs
concept_personus_scott_pelley, concept:personleadsorganization, concept_company_cnn__pbs
concept_actor_daniel_schorr, concept:worksfor, concept_televisionnetwork_cbs
concept_actor_daniel_schorr, concept:personleadsorganization, concept_televisionnetwork_cbs
concept_actor_daniel_schorr, concept:personleadsorganization, concept_company_cnn__pbs
concept_person_edward_r__murrow, concept:worksfor, concept_televisionnetwork_cbs
concept_person_edward_r__murrow, concept:agentcollaborateswithagent_inv, concept_televisionnetwork_cbs
concept_person_edward_r__murrow, concept:personleadsorganization, concept_televisionnetwork_cbs
concept_person_edward_r__murrow, concept:personleadsorganization, concept_company_cnn__pbs
concept_televisionnetwork_cbs, concept:organizationheadquarteredincity, concept_city_new_york
concept_televisionnetwork_cbs, concept:headquarteredin, concept_city_new_york
concept_televisionnetwork_cbs, concept:agentcollaborateswithagent, concept_personeurope_william_paley
concept_televisionnetwork_cbs, concept:topmemberoforganization_inv, concept_personeurope_william_paley
concept_company_cnn__pbs, concept:headquarteredin, concept_city_new_york
concept_company_cnn__pbs, concept:personbelongstoorganization_inv, concept_personeurope_william_paley
concept_nonprofitorganization_cbs_evening, concept:worksfor_inv, concept_journalist_walter_cronkite
concept_city_new_york, concept:organizationheadquarteredincity_inv, concept_televisionnetwork_cbs
concept_city_new_york, concept:headquarteredin_inv, concept_televisionnetwork_cbs
concept_city_new_york, concept:headquarteredin_inv, concept_company_cnn__pbs
concept_personeurope_william_paley, concept:agentcollaborateswithagent_inv, concept_televisionnetwork_cbs
concept_personeurope_william_paley, concept:topmemberoforganization, concept_televisionnetwork_cbs
concept_personeurope_william_paley, concept:personbelongstoorganization, concept_company_cnn__pbs
concept_personeurope_william_paley, concept:personleadsorganization, concept_company_cnn__pbs
\end{lstlisting}

\textbf{For the OrganizationHiredPerson task}

\begin{figure}[h]
  \hspace{-35pt}
  \includegraphics[width=1.2\textwidth]{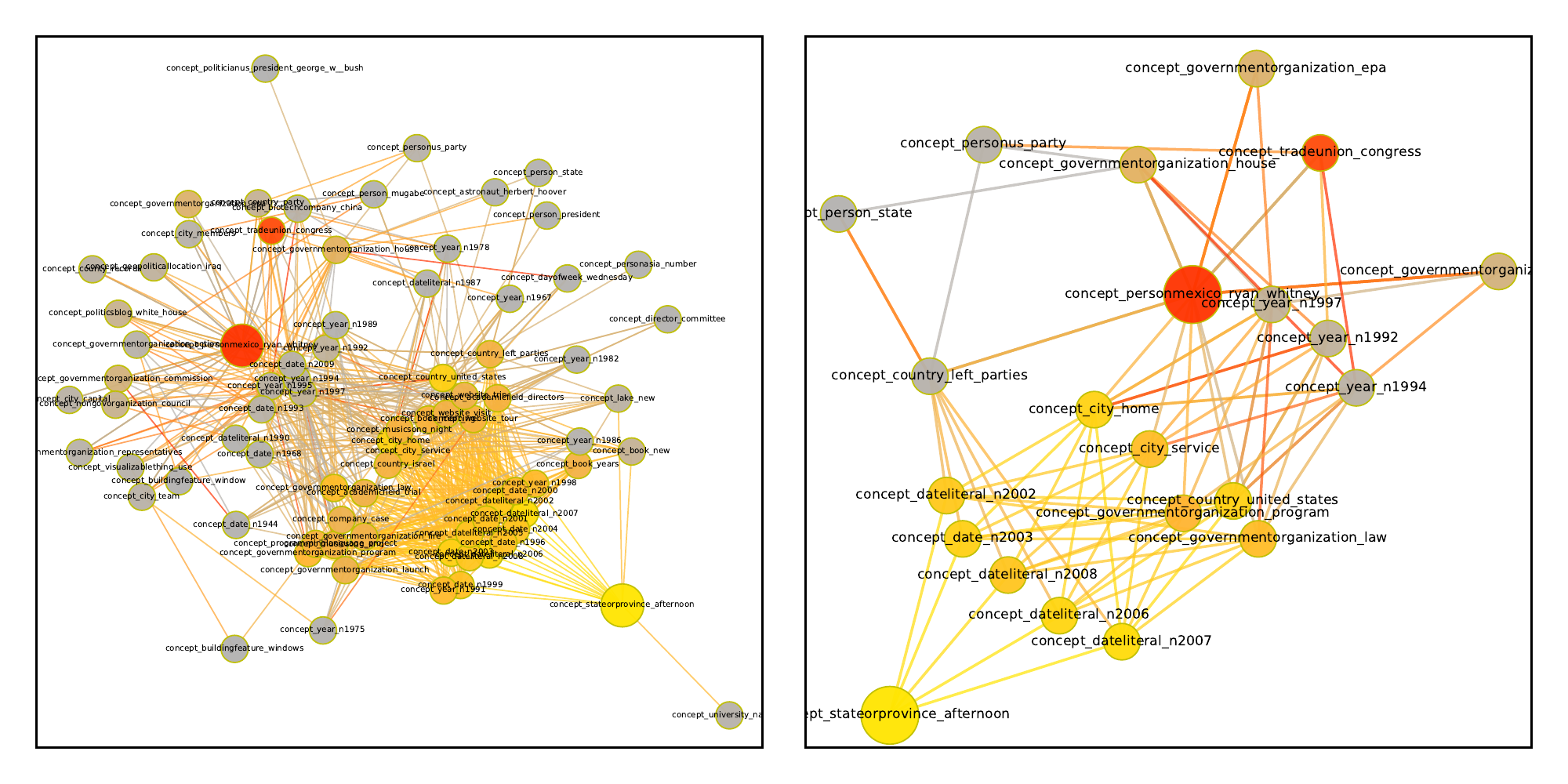}
  \caption{\textbf{OrganizationHiredPerson}. The head is \textit{concept\_stateorprovince\_afternoon}, the query relation is \textit{concept:organizationhiredperson}, and the desired tail is \textit{concept\_personmexico\_ryan\_whitney}. The left is a full subgraph derived with $max\_attended\_nodes\_per\_step = 20$, and the right is a further extracted subgraph from the left based on attention. The big yellow node represents the head, and the big red node represents the tail. Colors indicate how important a node is attended to in a local subgraph. Grey means less important, yellow means it is more attended during the early steps, and red means it is more attended when getting close to the final step.}
  \label{fig:}
\end{figure}

\begin{lstlisting}[basicstyle=\sffamily\scriptsize]
Query: (concept_stateorprovince_afternoon, concept:organizationhiredperson, concept_personmexico_ryan_whitney)

Selected key edges:
concept_stateorprovince_afternoon, concept:atdate, concept_dateliteral_n2007
concept_stateorprovince_afternoon, concept:atdate, concept_date_n2003
concept_stateorprovince_afternoon, concept:atdate, concept_dateliteral_n2006
concept_dateliteral_n2007, concept:atdate_inv, concept_country_united_states
concept_dateliteral_n2007, concept:atdate_inv, concept_city_home
concept_dateliteral_n2007, concept:atdate_inv, concept_city_service
concept_dateliteral_n2007, concept:atdate_inv, concept_country_left_parties
concept_date_n2003, concept:atdate_inv, concept_country_united_states
concept_date_n2003, concept:atdate_inv, concept_city_home
concept_date_n2003, concept:atdate_inv, concept_city_service
concept_date_n2003, concept:atdate_inv, concept_country_left_parties
concept_dateliteral_n2006, concept:atdate_inv, concept_country_united_states
concept_dateliteral_n2006, concept:atdate_inv, concept_city_home
concept_dateliteral_n2006, concept:atdate_inv, concept_city_service
concept_dateliteral_n2006, concept:atdate_inv, concept_country_left_parties
concept_country_united_states, concept:atdate, concept_year_n1992
concept_country_united_states, concept:atdate, concept_year_n1997
concept_country_united_states, concept:organizationhiredperson, concept_personmexico_ryan_whitney
concept_city_home, concept:atdate, concept_year_n1992
concept_city_home, concept:atdate, concept_year_n1997
concept_city_home, concept:organizationhiredperson, concept_personmexico_ryan_whitney
concept_city_service, concept:atdate, concept_year_n1992
concept_city_service, concept:atdate, concept_year_n1997
concept_city_service, concept:organizationhiredperson, concept_personmexico_ryan_whitney
concept_country_left_parties, concept:worksfor_inv, concept_personmexico_ryan_whitney
concept_country_left_parties, concept:organizationhiredperson, concept_personmexico_ryan_whitney
concept_year_n1992, concept:atdate_inv, concept_governmentorganization_house
concept_year_n1992, concept:atdate_inv, concept_country_united_states
concept_year_n1992, concept:atdate_inv, concept_city_home
concept_year_n1992, concept:atdate_inv, concept_tradeunion_congress
concept_year_n1997, concept:atdate_inv, concept_governmentorganization_house
concept_year_n1997, concept:atdate_inv, concept_country_united_states
concept_year_n1997, concept:atdate_inv, concept_city_home
concept_personmexico_ryan_whitney, concept:worksfor, concept_governmentorganization_house
concept_personmexico_ryan_whitney, concept:worksfor, concept_tradeunion_congress
concept_personmexico_ryan_whitney, concept:worksfor, concept_country_left_parties
concept_governmentorganization_house, concept:personbelongstoorganization_inv, concept_personus_party
concept_governmentorganization_house, concept:worksfor_inv, concept_personmexico_ryan_whitney
concept_governmentorganization_house, concept:organizationhiredperson, concept_personmexico_ryan_whitney
concept_tradeunion_congress, concept:organizationhiredperson, concept_personus_party
concept_tradeunion_congress, concept:worksfor_inv, concept_personmexico_ryan_whitney
concept_tradeunion_congress, concept:organizationhiredperson, concept_personmexico_ryan_whitney
concept_country_left_parties, concept:organizationhiredperson, concept_personus_party
\end{lstlisting}

\textbf{For the AgentBelongsToOrganization task}

\begin{figure}[h]
  \hspace{-35pt}
  \includegraphics[width=1.2\textwidth]{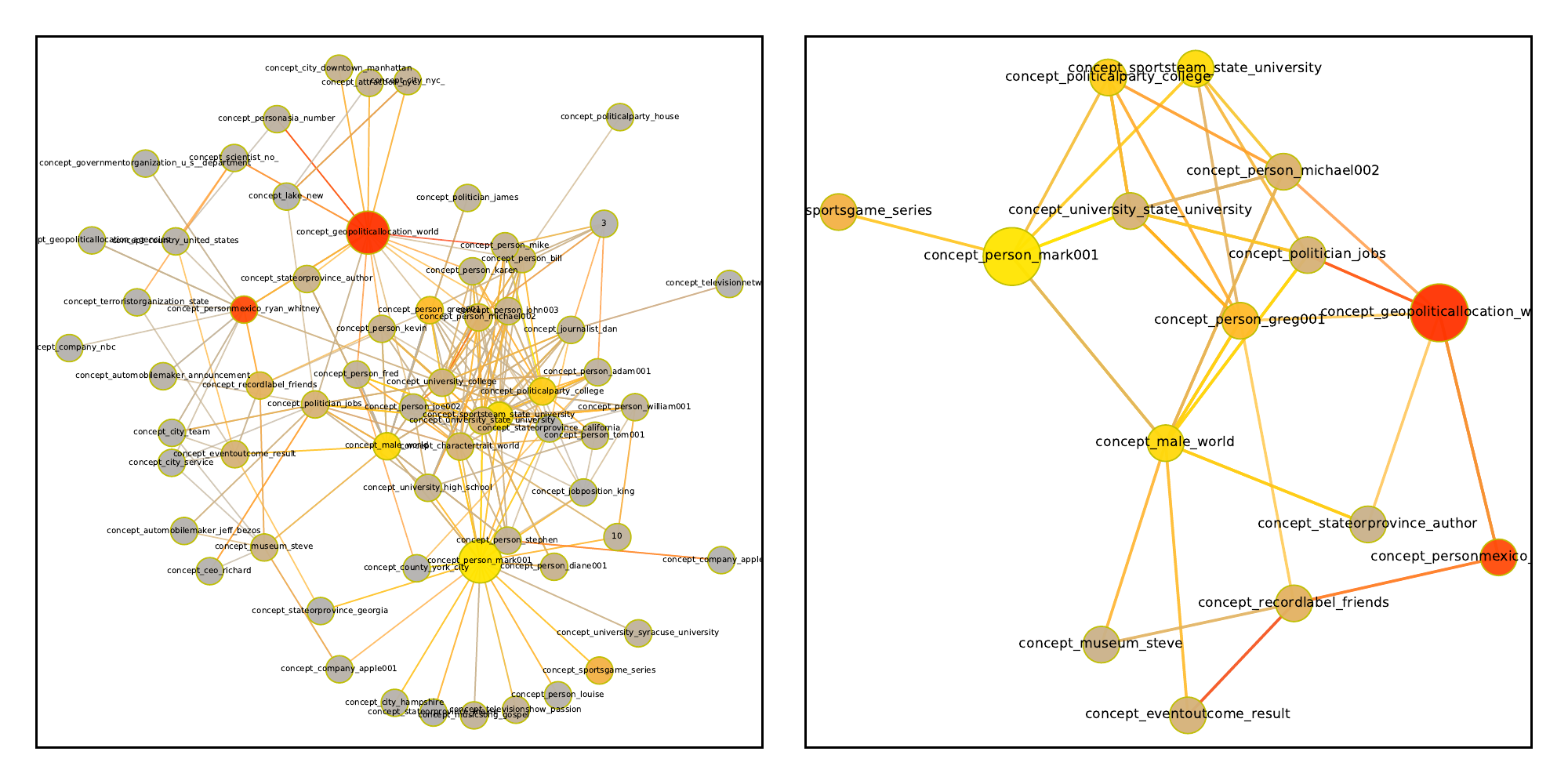}
  \caption{\textbf{AgentBelongsToOrganization}. The head is \textit{concept\_person\_mark001}, the query relation is \textit{concept:agentbelongstoorganization}, and the desired tail is \textit{concept\_geopoliticallocation\_world}. The left is a full subgraph derived with $max\_attended\_nodes\_per\_step = 20$, and the right is a further extracted subgraph from the left based on attention. The big yellow node represents the head, and the big red node represents the tail. Colors indicate how important a node is attended to in a local subgraph. Grey means less important, yellow means it is more attended during the early steps, and red means it is more attended when getting close to the final step.}
  \label{fig:}
\end{figure}

\begin{lstlisting}[basicstyle=\sffamily\scriptsize]
Query: (concept_person_mark001, concept:agentbelongstoorganization, concept_geopoliticallocation_world)

Selected key edges:
concept_person_mark001, concept:personbelongstoorganization, concept_sportsteam_state_university
concept_person_mark001, concept:agentcollaborateswithagent, concept_male_world
concept_person_mark001, concept:agentcollaborateswithagent_inv, concept_male_world
concept_person_mark001, concept:personbelongstoorganization, concept_politicalparty_college
concept_sportsteam_state_university, concept:personbelongstoorganization_inv, concept_politician_jobs
concept_sportsteam_state_university, concept:personbelongstoorganization_inv, concept_person_mark001
concept_sportsteam_state_university, concept:personbelongstoorganization_inv, concept_person_greg001
concept_sportsteam_state_university, concept:personbelongstoorganization_inv, concept_person_michael002
concept_male_world, concept:agentcollaborateswithagent, concept_politician_jobs
concept_male_world, concept:agentcollaborateswithagent_inv, concept_politician_jobs
concept_male_world, concept:agentcollaborateswithagent, concept_person_mark001
concept_male_world, concept:agentcollaborateswithagent_inv, concept_person_mark001
concept_male_world, concept:agentcollaborateswithagent, concept_person_greg001
concept_male_world, concept:agentcollaborateswithagent_inv, concept_person_greg001
concept_male_world, concept:agentcontrols, concept_person_greg001
concept_male_world, concept:agentcollaborateswithagent, concept_person_michael002
concept_male_world, concept:agentcollaborateswithagent_inv, concept_person_michael002
concept_politicalparty_college, concept:personbelongstoorganization_inv, concept_person_mark001
concept_politicalparty_college, concept:personbelongstoorganization_inv, concept_person_greg001
concept_politicalparty_college, concept:personbelongstoorganization_inv, concept_person_michael002
concept_politician_jobs, concept:personbelongstoorganization, concept_sportsteam_state_university
concept_politician_jobs, concept:agentcollaborateswithagent, concept_male_world
concept_politician_jobs, concept:agentcollaborateswithagent_inv, concept_male_world
concept_politician_jobs, concept:worksfor, concept_geopoliticallocation_world
concept_person_greg001, concept:personbelongstoorganization, concept_sportsteam_state_university
concept_person_greg001, concept:agentcollaborateswithagent, concept_male_world
concept_person_greg001, concept:agentcollaborateswithagent_inv, concept_male_world
concept_person_greg001, concept:agentcontrols_inv, concept_male_world
concept_person_greg001, concept:agentbelongstoorganization, concept_geopoliticallocation_world
concept_person_greg001, concept:personbelongstoorganization, concept_politicalparty_college
concept_person_greg001, concept:agentbelongstoorganization, concept_recordlabel_friends
concept_person_michael002, concept:personbelongstoorganization, concept_sportsteam_state_university
concept_person_michael002, concept:agentcollaborateswithagent, concept_male_world
concept_person_michael002, concept:agentcollaborateswithagent_inv, concept_male_world
concept_person_michael002, concept:agentbelongstoorganization, concept_geopoliticallocation_world
concept_person_michael002, concept:personbelongstoorganization, concept_politicalparty_college
concept_geopoliticallocation_world, concept:worksfor_inv, concept_personmexico_ryan_whitney
concept_geopoliticallocation_world, concept:organizationhiredperson, concept_personmexico_ryan_whitney
concept_geopoliticallocation_world, concept:worksfor_inv, concept_politician_jobs
concept_recordlabel_friends, concept:organizationhiredperson, concept_personmexico_ryan_whitney
concept_personmexico_ryan_whitney, concept:worksfor, concept_geopoliticallocation_world
concept_personmexico_ryan_whitney, concept:organizationhiredperson_inv, concept_geopoliticallocation_world
concept_personmexico_ryan_whitney, concept:organizationhiredperson_inv, concept_recordlabel_friends
\end{lstlisting}

\textbf{For the TeamPlaysInLeague task}

\begin{figure}[h]
  \hspace{-35pt}
  \includegraphics[width=1.2\textwidth]{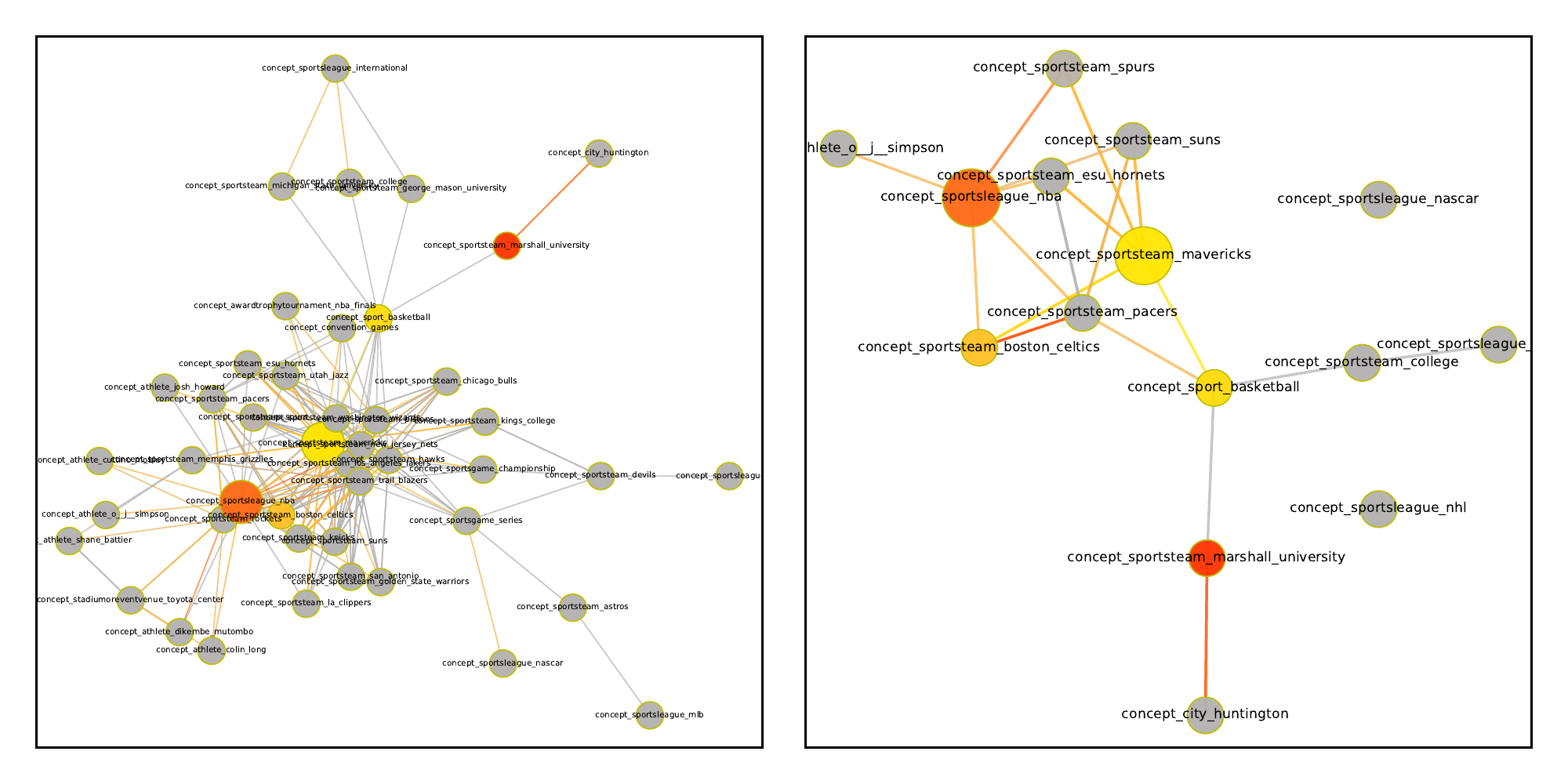}
  \caption{\textbf{TeamPlaysInLeague}. The head is \textit{concept\_sportsteam\_mavericks}, the query relation is \textit{concept:teamplaysinleague}, and the desired tail is \textit{concept\_sportsleague\_nba}. The left is a full subgraph derived with $max\_attended\_nodes\_per\_step = 20$, and the right is a further extracted subgraph from the left based on attention. The big yellow node represents the head, and the big red node represents the tail. Colors indicate how important a node is attended to in a local subgraph. Grey means less important, yellow means it is more attended during the early steps, and red means it is more attended when getting close to the final step.}
  \label{fig:}
\end{figure}

\begin{lstlisting}[basicstyle=\sffamily\scriptsize]
Query: (concept_sportsteam_mavericks, concept:teamplaysinleague, concept_sportsleague_nba)

Selected key edges:
concept_sportsteam_mavericks, concept:teamplayssport, concept_sport_basketball
concept_sportsteam_mavericks, concept:teamplaysagainstteam, concept_sportsteam_boston_celtics
concept_sportsteam_mavericks, concept:teamplaysagainstteam_inv, concept_sportsteam_boston_celtics
concept_sportsteam_mavericks, concept:teamplaysagainstteam, concept_sportsteam_spurs
concept_sportsteam_mavericks, concept:teamplaysagainstteam_inv, concept_sportsteam_spurs
concept_sport_basketball, concept:teamplayssport_inv, concept_sportsteam_college
concept_sport_basketball, concept:teamplayssport_inv, concept_sportsteam_marshall_university
concept_sportsteam_boston_celtics, concept:teamplaysinleague, concept_sportsleague_nba
concept_sportsteam_spurs, concept:teamplaysinleague, concept_sportsleague_nba
concept_sportsleague_nba, concept:agentcompeteswithagent, concept_sportsleague_nba
concept_sportsleague_nba, concept:agentcompeteswithagent_inv, concept_sportsleague_nba
concept_sportsteam_college, concept:teamplaysinleague, concept_sportsleague_international
\end{lstlisting}